\documentclass[conference]{IEEEtran}
\IEEEoverridecommandlockouts

\usepackage[utf8]{inputenc} 
\usepackage[T1]{fontenc}    
\usepackage{hyperref}       
\usepackage{url}            
\usepackage{booktabs}       
\usepackage{amsfonts}       
\usepackage{nicefrac}       
\usepackage{microtype}      
\usepackage{xcolor}         
\usepackage{hyperref}
\usepackage{url}
\usepackage{graphicx}
\usepackage{wrapfig}
\usepackage{booktabs}
\usepackage{capt-of}
\usepackage{multirow}
\usepackage[normalem]{ulem}
\usepackage{subcaption}  
\usepackage{comment}
\usepackage{orcidlink}

\usepackage{cite}
\usepackage{amsmath,amssymb,amsfonts}
\usepackage{algorithmic}
\usepackage{textcomp}
\def\BibTeX{{\rm B\kern-.05em{\sc i\kern-.025em b}\kern-.08em
    T\kern-.1667em\lower.7ex\hbox{E}\kern-.125emX}}

\begin{document}

\title{Adversarial Attacks on Downstream Weather Forecasting Models: Application to Tropical Cyclone Trajectory Prediction}

\author{\IEEEauthorblockN{Yue Deng\orcidlink{0000-0002-8263-1871}}
\IEEEauthorblockA{
\textit{Michigan State University}\\
East Lansing, MI 48824 USA \\
dengyue1@msu.edu}
\and
\IEEEauthorblockN{Francisco Santos\orcidlink{0000-0001-9929-5299}}
\IEEEauthorblockA{
\textit{Michigan State University}\\
East Lansing, MI 48824 USA \\
santosf3@msu.edu}
\and
\IEEEauthorblockN{Pang-Ning Tan\orcidlink{0000-0003-3205-0339}}
\IEEEauthorblockA{
Michigan State University\\
East Lansing, MI 48824 USA \\
ptan@msu.edu}
\and
\IEEEauthorblockN{Lifeng Luo\orcidlink{0000-0002-2829-7104}}
\IEEEauthorblockA{
\textit{Michigan State University}\\
East Lansing, MI 48824 USA \\
lluo@msu.edu}
}

\maketitle

\begin{abstract}
Deep learning–based weather forecasting (DLWF) models leverage past weather observations to generate future forecasts, supporting a wide range of downstream applications, including tropical cyclone (TC) prediction. In this paper, we investigate the vulnerability of these models to adversarial attacks, where subtle perturbations to the upstream forecasts can alter the downstream TC trajectory predictions. Crafting perturbed inputs to steer the downstream outputs toward attacker-specified trajectories is a challenging problem for several reasons. First, conventional TC detection systems are opaque, non-differentiable black boxes, making standard gradient-based attacks infeasible. Second, the extreme rarity of TC events leads to a severe class imbalance problem, making it difficult to develop attack methods that produce realistic-looking cyclone paths aligned with the attacker’s target trajectories. To overcome these limitations, we propose Cyc-Attack, a novel method for perturbing the forecasts of DLWF models to generate adversarial trajectories. The proposed method uses a differentiable surrogate model to approximate the TC detector's output, enabling the application of gradient-based attacks. Cyc-Attack also employs a skewness-aware loss function with a kernel dilation strategy to address the severe class imbalance problem. Distance-based gradient weighting and regularization strategies are also introduced to constrain the perturbations to eliminate unrealistic-looking TC paths. Our experiments show that Cyc-Attack achieves a higher true positive rate in matching the attacker’s target trajectories, along with lower false alarm rates and stealthier perturbations than conventional attack methods. We further demonstrate that the adversarial upstream forecasts generated by Cyc-Attack can be leveraged as initial guidance to train an end-to-end model for learning adversarial input samples that would induce the attacker’s target trajectories. Our code and supplementary materials are available at \url{https://github.com/dengy0111/Cyc-Attack}.
\end{abstract}


\section{Introduction}
\label{sec:Introduction}

Recent progress in deep learning–based weather forecasting (DLWF) models~\cite{gao2022earthformer,lin2022conditional,lam2023learning,bi2023accurate} has yielded notable gains in predictive accuracy, often surpassing conventional numerical models. However, the DLWF models are vulnerable to adversarial attacks~\cite{deng2025fable, imgrund2025adversarial, arif2025forecasting}, in which subtle manipulations of the input data can lead to significant changes in the output forecasts at targeted locations. The potential for such disruptions is demonstrated 
by the 2014 cyber intrusion into NOAA systems that disrupted satellite data used by numerical weather prediction models\footnote{\url{https://time.com/3581563/chinese-hackers-noaa-breach-weather/}}. More recently, French authorities investigated suspected tampering of temperature sensors at the Paris Charles de Gaulle Airport after unusual temperature spikes coincided with profitable weather bets on Polymarket\footnote{\url{https://decrypt.co/365276}.}. 


\begin{figure*}[tb]
     \centering
     \begin{subfigure}{0.52\textwidth}
         \centering
        \includegraphics[width=\linewidth]{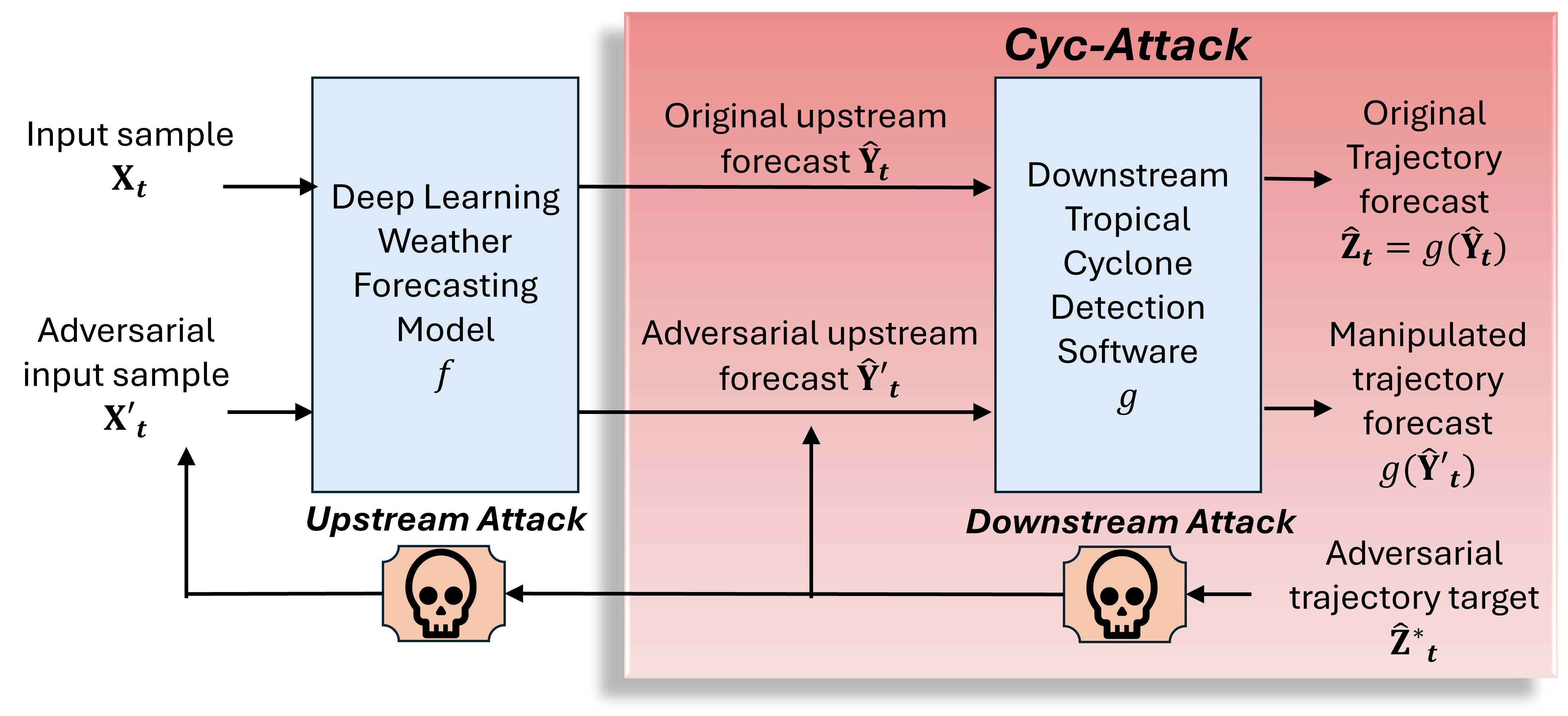}  
    \caption{Threat model for downstream prediction.}  
    \label{fig:formulation}
    \end{subfigure}
    \hspace{0.01\textwidth}
    \begin{subfigure}{0.45\textwidth}
     \centering
    \includegraphics[width=\linewidth]{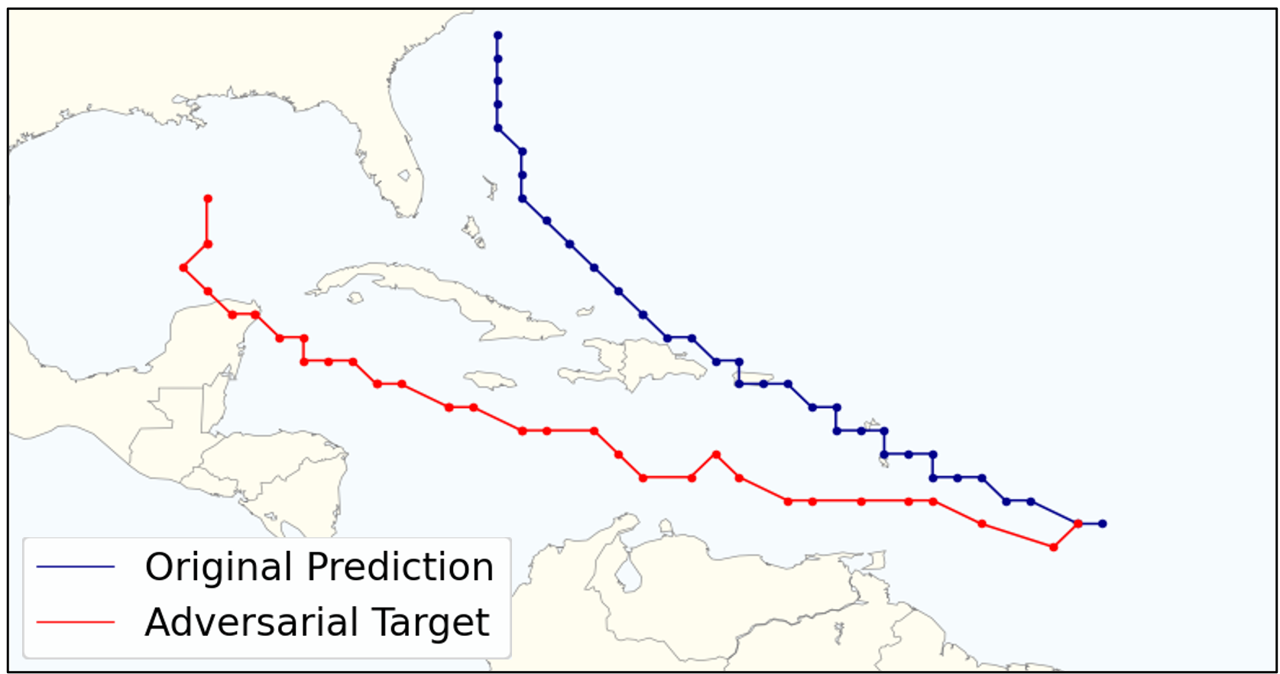}  
    \caption{Manipulated trajectory of hurricane Maria.}
    \label{fig:hurricane_track}
     \end{subfigure}
     \caption{Adversarial attack on downstream tropical cyclone (TC) trajectory prediction. The figure on the right shows adversarial manipulation of hurricane Maria's projected trajectory, generated by the \textit{TempestExtremes} software from the 10-day output forecast of the \emph{GraphCast} model, steering its original forecasted path (shown as \textcolor{blue}{blue} line) towards a targeted region (shown as \textcolor{red}{red} line).
     }
     \label{fig:DLWF_attack}
\end{figure*}

This paper investigates the feasibility of constructing adversarial weather forecasts for downstream applications. As weather forecasts are often used to inform planning and decision-making, \textbf{\emph{adversarial targets are more naturally defined at the downstream level}}, for example, as an altered tropical cyclone (TC) path rather than as a perturbed set of meteorological variables. This allows an attacker to specify the desired target without requiring considerable domain knowledge. 
We focus on TC trajectory prediction~\cite{wang2022review} as our downstream application given its significant socio-economic impacts, as evidenced by the more than \$2.9 trillion in damages caused by TCs in the United States since 1980\footnote{\url{https://coast.noaa.gov/states/fast-facts/hurricane-costs.html}}, as well as the limited prior research in this critical domain. Figure~\ref{fig:hurricane_track} illustrates an example scenario in which an adversary manipulates the projected trajectory of Hurricane \textit{Maria}, diverting it from its original path towards a region with extensive oil refinery infrastructure, with the intent of disrupting the energy market and profiting from the market instability. Such an attack, if successful, could have left the densely populated U.S. East Coast under-prepared while causing costly planning and resource misallocation at the targeted region. This scenario highlights the necessity of investigating the feasibility of adversarial attacks on DLWF models for TC trajectory predictions.

While adversarial attack on DLWF models is an active research area, their implications for downstream tasks such as TC prediction remain largely unexplored, with several key challenges yet to be resolved. 
In particular, the black-box nature of most TC detection systems~\cite{ullrich2021tempestextremes,perez2024cytrack,yan2023trophy} renders conventional gradient-based attack methods, such as FGSM~\cite{goodfellow2014explaining}, PGD~\cite{madry2017towards}, or CW Attack~\cite{carlini2017towards}, infeasible. One way to overcome this limitation is to pre-train a surrogate model to replicate the output of the TC detection system, thereby enabling the application of gradient-based methods. However, training a robust surrogate model that faithfully captures the behavior of a TC detection system remains a challenging task in its own right. First, the surrogate model must address the severe class imbalance problem~\cite{johnson2019survey}, as the proportion of locations experiencing a TC event at any given time step is extremely small ($< 0.01\%$) compared to that of unaffected locations. Failure to address this challenge may compromise the quality of the generated adversarial samples.  
Specifically, incorrect predictions made by the surrogate model may cause gradient-based optimization methods to terminate prematurely by falsely indicating that the target objective has been achieved. Another challenge is to ensure that the adversarial attack remains stealthy by keeping the perturbations to the original inputs small while preserving realisticness of the resulting TC trajectories. In particular, the generated adversarial trajectories should be free of conspicuous artifacts, such as zigzag paths, that could make the adversarial attack easily detectable.

To address these challenges, we introduce {\textit{Cyc-Attack}}, a novel method for perturbing the upstream forecasts of DLWF models to generate adversarial TC trajectories (see Figure~\ref{fig:formulation}). {\textit{Cyc-Attack}} employs a differentiable surrogate model to approximate the output of TC detection systems, enabling efficient generation of adversarial upstream forecasts through gradient-based optimization. This surrogate-based approach is substantially more query-efficient than gradient-free methods, such as zeroth-order optimization~\cite{berahas2022theoretical,lian2016comprehensive,nesterov2017random,chen2024deepzero}, which typically require a large number of queries to the black-box system to generate adversarial samples. To address the severe class imbalance issue when training the surrogate model, we employ a skewness-aware loss function along with a kernel dilation strategy, guiding the surrogate model to better predict the true TC locations. The same strategy is also applied during adversarial forecast generation. To further mitigate the effects of inaccuracies in the surrogate model during adversarial forecast generation, we assign greater weights to the gradient contributions of incorrectly classified samples in the loss function, to prevent premature termination of the optimization process. We also employ distance-based gradient weighting and regularization to constrain deviations from the original DLWF upstream forecast, ensuring that the perturbed forecasts remain realistic and free of irregular trajectory distortions when processed by the TC detection system. 

Finally, our experimental results demonstrate that {\textit{Cyc-Attack}} achieves a higher true positive rate in steering the TC detection system toward the attacker's target trajectories, while simultaneously yielding lower false alarm rates and less detectable perturbations than conventional attack methods. Furthermore, we show that the adversarial upstream forecasts generated by {\textit{Cyc-Attack}} can serve as effective initial guidance for training an end-to-end model to generate adversarial input samples that induce the attacker's target trajectories.

\section{Related Work}
\label{sec:Related Work}

Modern weather forecasting systems operate on high-dimensional data characterized by multiscale dynamics and strong cross-variable couplings. 
Beyond conventional numerical weather prediction (NWP) models~\cite{cote1998operational,skamarock2008description,hersbach2020era5} that solve the governing physical equations to produce forecasts, recent DLWF methods~\cite{lam2023learning,gao2022earthformer,bi2023accurate,price2025probabilistic,bonev2025fourcastnet,couairon2026archesweathergen,rasp2026weathernext,lang2026aifs} have emerged as promising alternatives. 
These models learn complex spatiotemporal patterns from data and have demonstrated improved prediction accuracy compared to conventional NWP models. The forecasted weather can be used for various downstream applications, including TC detection and tracking. For example, \textit{TempestExtremes}~\cite{ullrich2021tempestextremes} is an open-source tool that employs expert-defined rules to detect TC tracks from forecasted meteorological variables such as wind speed, mean sea level pressure (MSLP), and geopotential thickness. It first identifies TC candidate nodes at each time step based on local minima in mean sea level pressure with warm-core signatures and links together those nodes satisfying certain criteria (e.g., wind speed beyond a pre-specified threshold) into trajectories. Other automated TC trajectory detection tools include \textit{CyTrack}~\cite{perez2024cytrack} and \textit{TROPHY}~\cite{yan2023trophy}. 

Recent studies have shown that DLWF models are vulnerable to adversarial attacks. For example, Chichifoi et al.~\cite{chichifoi2025evaluating} showed that compromising only a small subset of clients in a federated learning setting can lead to substantial distortions in temperature forecasts across large spatial regions. Imgrund et al.~\cite{imgrund2025adversarial} proposed adversarial attacks on diffusion-based DLWF models~\cite{price2025probabilistic} by perturbing regional observations to amplify forecast errors and induce spurious extreme-event predictions. Deng et al.~\cite{deng2025fable} developed localized targeted attacks that steer forecasts at selected locations toward attacker-specified values while preserving closeness and geospatio-temporal realisticness. However, none of these studies investigates adversarial attacks from the perspective of downstream tasks, such as manipulating TC trajectories, which is the focus of this paper. 

Adversarial attacks can be categorized by attack objective~\cite{heinrich2024targeted} as \textit{untargeted}, \textit{semi-targeted}, or \textit{targeted}, or by attacker's knowledge of the model~\cite{liu2022practical} as \textit{white-box} (full access to model architecture and parameters), \textit{gray-box} (partial knowledge), or \textit{black-box} (query access only). Representative white-box methods include FGSM~\cite{goodfellow2014explaining}, PGD~\cite{madry2017towards}, and CW~\cite{carlini2017towards}. Black-box attacks generally adopt either surrogate models~\cite{papernot2016transfer,papernot2017practical}, which enable gradient-based optimization through a proxy model, or zeroth-order (ZO) optimization~\cite{chen2024deepzero}, which estimates gradient directions using queries to the target model. Since our work focuses on attacking a black-box downstream TC detector, we adopt a surrogate-based approach, avoiding the prohibitive computational cost of estimating gradients through black-box queries via ZO optimization. 

\section{Adversarial Threat Model}
\label{sec:Problem statement}

Let $\mathbf{X} \in\mathbb{R}^{\tau\times d\times r\times c}$ be a set of current and past weather observations for $r \times c$ gridded locations, where $\tau$ is the window size and $d$ is the number of weather variables. We consider a DLWF model $f$ that generates forecasts of future weather conditions from $\mathbf{X}$, i.e., $\mathbf{\hat{Y}} = f(\mathbf{X}) \in\mathbb{R}^{\beta\times d\times r\times c}$, where $\beta$ is the forecast horizon. The upstream forecast $\mathbf{\hat{Y}}$ will be provided to a downstream detector $g$ to generate a projected TC trajectory, $\hat{\mathbf{Z}} = g(\mathbf{\hat{Y}}) \in \{0,1\}^{\beta \times r \times c}$, where $\hat{\mathbf{Z}}_{tij}=1$ if location $(i,j)$ is experiencing a TC event at time $t$. Otherwise, $\hat{\mathbf{Z}}_{tij}=0$. 

The attacker's goal is to alter the TC trajectory from its original forecast $\hat{\mathbf{Z}} = g(f(\mathbf{X}))$ to the adversarial target, $\hat{\mathbf{Z}}^*\in\{0,1\}^{\beta\times r\times c}$. The adversarial attack can be conducted at two attack points: one at the input of the upstream DLWF model and the other at the output of the upstream model. We termed the former an \textbf{upstream attack} and the latter as \textbf{downstream attack}, as illustrated in Figure~\ref{fig:formulation}. For the downstream attack, the attacker aims to generate an adversarial upstream forecast $\hat{\mathbf{Y}}'$ so that the detector $g$ will produce a TC trajectory $g(\hat{\mathbf{Y}}')$ that closely matches the desired target $\hat{\mathbf{Z}}^*$. The 
adversarial upstream forecast can also be used to perform an upstream attack by using $\hat{\mathbf{Y}}'$ to guide the construction of an adversarial input $\mathbf{X}'$ such that $g(f(\mathbf{X'}))$ closely matches $\hat{\mathbf{Z}}^*$.


Our threat model assumes that the defender's goal is to identify whether the input to the upstream DLWF model has been manipulated. To this end, the defender trains an anomaly detector using a set of clean historical weather observations. Given a potentially manipulated input $\hat{\mathbf{X}}$, the detector evaluates its likelihood under the learned distribution, thereby estimating the probability that the input sample is consistent with the historical observations.


We assume that the TC detector $g$ operates as a black box, whereas the upstream DLWF model $f$ is treated as a white box. This assumption is practical, as the source code and pretrained checkpoints of most widely used DLWF models are 
publicly available~\cite{lam2023learning,price2025probabilistic,bonev2025fourcastnet,couairon2026archesweathergen,lang2026aifs}, providing the attacker with access to the upstream model's internal architecture and parameters. In this paper, we employ \textit{GraphCast}~\cite{lam2023learning}, a DLWF model developed by Google DeepMind, as the forecasting model with \textit{TempestExtremes}~\cite{ullrich2021tempestextremes} as the downstream TC trajectory detector. However, due to its model-agnostic nature, our framework is applicable to other DLWF models and TC detectors. 

To enable gradient-based attacks, we approximate the black-box detector $g$ with a differentiable surrogate model $\tilde{g}$. Given an upstream forecast $\hat{\mathbf{Y}}$, the surrogate model produces an output $\hat{\mathbf{P}} = \tilde{g}(\hat{\mathbf{Y}})\in[0,1]^{\beta\times r\times c}$, where $\hat{P}_{tij}$ represents the predicted probability that location $(i,j)$ experiences a TC event at time $t$. These probabilities can be converted into binary predictions using a threshold of $0.5$, i.e., a TC event is predicted if $\hat{P}_{tij}\geq 0.5$ and no event is predicted otherwise. The surrogate model $\tilde{g}$ is pre-trained to approximate the TC trajectories produced by the black-box detector, i.e., $\tilde{g}(\hat{\mathbf{Y}})\approx g(\hat{\mathbf{Y}})$. This approximation enables the application of gradient-based downstream attack algorithms. Specifically, given a target TC trajectory, $\hat{\mathbf{Z}}^*$, we apply a downstream attack algorithm $\mathcal{A}_d$ to the surrogate model to generate an adversarial upstream forecast: $\hat{\mathbf{Y}}' = \mathcal{A}_d(\tilde{g}, \hat{\mathbf{Y}}, \hat{\mathbf{Z}}^*)$. This adversarial forecast serves as an intermediate target for training an end-to-end upstream attack algorithm, $\mathcal{A}_u$. Specifically, $\mathcal{A}_u$ learns to generate an adversarial input $\mathbf{X}'$ such that the resulting upstream forecast produces the desired TC trajectory when processed by the original black-box detector, i.e., $g(f(\mathbf{X'})) \approx \hat{\mathbf{Z}}^*$, where $\mathbf{X}' =  \mathcal{A}_u(f, \tilde{g}, \mathbf{X}, \hat{\mathbf{Y}}', \hat{\mathbf{Z}}^*)$.

\section{Cyc-Attack: Downstream Attack on TC Trajectory Predictions}
\label{sec:Methodology}

{\textit{Cyc-Attack}} consists of 2 main components: (1) a differentiable surrogate model trained to approximate a TC detection system's output and (2) an adversarial forecast generation module. 
Given an original forecast $\hat{\mathbf{Y}}$, {\textit{Cyc-Attack}} creates an adversarial upstream forecast $\hat{\mathbf{Y}}'$, such that the TC detector outputs a trajectory closely aligned with the adversary’s target, $\hat{\mathbf{Z}}^*$, that is, $g(\hat{\mathbf{Y}}')\approx \hat{\mathbf{Z}}^*$. 

\subsection{Surrogate Model Pre-training}
\label{sec:Surrogate Model Pre-training}

To pre-train the surrogate model, we construct a training dataset of the form 
$\{(\mathbf{\hat{Y}}^{(i)}, \hat{\mathbf{Z}}^{(i)})\}_{i=1}^N$, where $\mathbf{\hat{Y}}^{(i)}$ denotes the upstream forecasts and $\hat{\mathbf{Z}}^{(i)}$ represents the corresponding maps of TC trajectory locations for hurricane $i$. Each training pair is generated by passing a historical weather input $\mathbf{X}^{(i)}$  through the DLWF model to obtain the upstream forecast $\hat{\mathbf{Y}}^{(i)}$. The forecast is then processed by the black-box TC detector $g$ to produce the corresponding TC trajectory map $\hat{\mathbf{Z}}^{(i)}$. For brevity, we omit the superscript $i$ throughout the remainder of this paper. Our pre-training framework offers the following advantages: (1) it trains the surrogate model to closely approximate the outputs of \textit{TempestExtremes} to facilitate gradient-based optimization for adversarial forecast generation; (2) it enables the surrogate model to handle different forecast horizons without requiring retraining; and (3) it supports the use of any deep image segmentation architectures, such as \textit{DeepLabV3+}~\cite{chen2018encoder}, as the surrogate model. 

The surrogate model $\tilde{g}$ is trained to map the upstream forecast at each time step $t$, $\hat{\mathbf{Y}}_t \in \mathbb{R}^{d\times r \times c}$, to its corresponding TC location map, $\hat{\mathbf{Z}}_t \in \mathbb{R}^{r \times c}$. Pre-training the surrogate model requires addressing the severe class imbalance issue~\cite{johnson2019survey} since TC-affected locations at each time step $t$ are substantially fewer than non-TC locations ($< 0.01\%$). Without appropriate mitigation measures, the surrogate model may become biased toward the majority class (i.e., non-TC locations), leading to an increased false negative rate. To address this issue, we train the surrogate model to identify spatially dilated regions surrounding the TC-affected locations, rather than performing strict pixel-wise binary classification in which each location  $(i,j)$ is labeled as either TC-affected ($\hat{Z}_{tij}=1$) or non-TC-affected ($\hat{Z}_{tij}=0$) at any given time step $t$. For each TC-affected location $(i,j)$, we define its spatial neighborhood as $\Omega_{ij} = \{(p,q): (p-i)^2+(q-j)^2 \le R_s^2\}$, where $R_s$ denotes the spatial dilation radius of the neighborhood. The dilated TC region data, denoted as $\mathbf{\hat{Z}}^{D}$, are generated from the original TC location data $\mathbf{\hat{Z}}$ using a truncated Gaussian kernel $K$: 
\begin{equation}
\forall (p,q) \in \Omega_{ij}: \hat{Z}^{D}_{tpq}
= 
\max\!\left\{
\hat{Z}_{tpq},\;\hat{Z}_{tij} K_{\sigma,R_S}(p-i,\;q-j)
\right\},
\label{eq:surrogate_dilation}
\end{equation}
where $$K_{\sigma,R_S}(u,v)=\begin{cases}
    \exp\!\big(-\tfrac{u^2+v^2}{2\sigma^2}\big) & \textrm{if $\sqrt{u^2+v^2}\le R_S$}\\
    0 & \textrm{otherwise}
\end{cases}$$ Here, $\sigma$ denotes the kernel width parameter. Intuitively, this dilation 
expands the TC-affected regions to include nearby locations, producing soft labels $\hat{\mathbf{Z}}^D \in [0,1]^{\beta \times r \times c}$ instead of binary labels in $\{0,1\}$. When a location falls within multiple neighborhoods (e.g., due to its proximity to two distinct TCs), we assign it the maximum soft-label value induced by these neighborhoods. This strategy ensures that each location retains the strongest local TC influence rather than having its label diluted by weaker contributions from more distant TC centers.

To further alleviate the class imbalance issue, the surrogate model is trained to minimize the following skewness-aware loss, which is a variation of the focal loss~\cite{lin2017focal} used in object detection: 
\begin{align}
\mathcal{L}_{\text{surrogate}} 
= -\frac{1}{\beta}\sum_{t=1}^{\beta}\sum_{p=1}^{r}\sum_{q=1}^{c}\Big[(1 - \hat{P}_{tpq})^{2}\,\hat{Z}^{D}_{tpq}\log (\hat{P}_{tpq}) \nonumber\\
+ \hat{P}_{tpq}^{2}(1 - \hat{Z}^{D}_{tpq})\log (1 - \hat{P}_{tpq})\Big],
\label{eq:dilated_hurricane_locations}
\end{align}
where $\hat{P}_{tpq}$ is the probability predicted by the surrogate $\tilde{g}$ that location $(p,q)$ at time $t$ is in a TC-affected region and $\hat{Z}^{D}_{tpq}$ is the dilated soft label. 
Unlike the standard focal loss, which down-weights easy-to-classify examples with large $\hat{P}_{tpq}$, our surrogate loss reweights the loss according to the soft labels defined by the dilated regions, $\hat{Z}^{D}_{tpq}$, to promote accurate predictions in and around the targeted regions.
Our surrogate model is trained to minimize the loss in Equation \eqref{eq:dilated_hurricane_locations} using the Adam optimizer~\cite{kingma2014adam}.

As will be shown in our experiments (see Table~\ref{tab:diff_R_surrogate_rates} and Figure~\ref{fig:surrogate_different_radius}), the standard focal loss without kernel dilation leads to numerous false-positive predictions of isolated TC events across the globe. This behavior stems from the design of focal loss, which addresses class imbalance by assigning greater weight to hard-to-classify examples~\cite{lin2017focal}. In our setting, however, each TC event is typically represented by only one or two pixels with $\hat{Z}_{tij}=1$ at each time step (depending on the number of hurricanes present), resulting in extremely sparse positive supervision. Consequently, the model tends to overemphasize these isolated positive pixels, increasing the likelihood of false-positive predictions at other locations. To address this issue, kernel dilation expands each TC location into a local region, assigning soft labels to neighboring pixels to alleviate the imbalance and provide spatial context. Since the ultimate objective remains to identify the precise one-pixel TC location, the soft-label values progressively decrease with distance from the center, emphasizing the true TC location while providing weaker supervision to surrounding pixels.

\subsection{Adversarial Weather Forecast Generation}
\label{sec:Adversarial Attack Implementation}

The pre-trained surrogate model $\tilde{g}$ will be used to 
generate an adversarial upstream forecast $\hat{\mathbf{Y}}'_t$ for each time step $t$ by perturbing the original forecast $\hat{\mathbf{Y}}_t$ in such a way that $g(\hat{\mathbf{Y}}'_t)\approx\hat{\mathbf{Z}}^*_t$. A standard approach for generating adversarial samples is to apply  
Projected Gradient Descent (PGD)~\cite{madry2017towards}, a widely-used gradient-based attack method. Given a differentiable loss $\mathcal{L}(\tilde{g}(\hat{\mathbf{Y}}_t), \hat{\mathbf{Z}}_t^*)$, PGD iteratively updates the upstream forecast $\hat{\mathbf{Y}}_t$ in the direction of the negative loss gradient to minimize the loss: $\hat{\mathbf{Y}}_t' \leftarrow \hat{\mathbf{Y}}_t'-\eta\,\mathrm{sign}(\nabla_{\hat{\mathbf{Y}}_t}\mathcal{L})$, where $\hat{\mathbf{Y}}_t'$ is initially set to $\hat{\mathbf{Y}}_t$ and $\eta$ is the step size.  Following each update, the perturbed forecast is projected onto an $\ell_\infty$-bounded forecast centered at the original forecast $\hat{\mathbf{Y}}_t$ to constrain the magnitude of the perturbation, i.e., $\|\hat{\mathbf{Y}}'_t-\hat{\mathbf{Y}}_t\|_\infty \le \delta$, where $\delta$ denotes the maximum allowable perturbation magnitude. 

However, the standard PGD attack method is insufficient for our downstream attack for several reasons: 

\textbf{First}, we observe that using $\hat{\mathbf{Z}}^*_t$ directly as the adversarial target may cause the standard PGD algorithm to generate a perturbed upstream forecast $\hat{\mathbf{Y}}'_t$ that, when processed by the surrogate model $\tilde{g}$, yields trajectories with a substantial number of false positive TC  locations dispersed across the map, making the attack easier to be detected. This issue parallels the class imbalance problem described in the previous section. To address this concern, we apply the kernel dilation strategy described in Equation~(\ref{eq:surrogate_dilation}), with dilation radius $R_A$, to construct a dilated adversarial downstream target $\hat{\mathbf{Z}}_t^{*D}\in[0,1]^{r\times c}$ from $\hat{\mathbf{Z}}^*_t$.

\textbf{Second}, as the surrogate model only approximates the output of the TC detector, its prediction errors can misguide the adversarial forecast generation process, causing the PGD algorithm to terminate prematurely.  For example, suppose the attacker specifies a target location that is originally predicted as non-TC-affected but wishes it to be classified as TC-affected, i.e., $\hat{Z}^*_{tij}=1$ while $\hat{Z}_{tij}=0$. Due to the approximation errors, the surrogate model may incorrectly classify the location as TC-affected, i.e., $\tilde{g}(\hat{\mathbf{Y}}_{tij}) =1 \neq \hat{Z}_{tij}$. As a result, the PGD algorithm will refrain from perturbing the upstream forecast at that location, as it falsely concludes that the attack objective has been achieved. A similar problem arises when 
$\hat{Z}^*_{tij}=0$ but $\hat{Z}_{tij}=1$. 
To mitigate this issue, we introduce a calibration mask $\mathbf{M}\in\{0,1\}^{\beta\times r\times c}$ that identifies the set of locations $\{(i,j)\}$ where the surrogate model already satisfies the attack objective due to its prediction errors. Specifically, the mask assigns a value of 1 to  location ($i,j$) at time $t$ if the adversarial target differs from the original prediction, yet the surrogate predicts the location as having the target value: 
\begin{equation}
M_{tij}=\mathbf{1}\!\big(\hat{Z}^*_{tij}\neq\hat{Z}_{tij}\big)\mathbf{1}\!\big(\tilde{g}(\hat{\mathbf{Y}}_{tij})=\hat{Z}'_{tij}\big),
\end{equation}
where $\mathbf{1}(\cdot)$ denotes the indicator function, whose output is 1 if its argument is true and 0 otherwise. As described below, the calibration mask prevents early stopping of the optimization process, ensuring that the perturbations continue to be applied even when the surrogate’s initial prediction already matches the adversarial target.

\textbf{Third}, while kernel dilation reduces scattered false positive locations, it expands the target region, thereby increasing the number of candidate locations considered by the TC detector to form a TC trajectory. This causes the detector to produce trajectories with pronounced zigzag patterns as shown in Figure~\ref{fig:adv_attack_different_radius}, making them unrealistic and easily detectable. To overcome this issue, we constrain the perturbation on $\hat{\mathbf{Y}}$ by (1) down-weighting the gradient update for locations far away from the original and targeted paths and (2) up-weighting the penalty for misclassifying such locations in the loss function. Specifically, let $\mathcal{S}_t = \{(i,j): \hat{Z}^*_{tij}=1 \ \textrm{or} \ \hat{Z}_{tij}=1\}$ be the set of locations in the regions associated with the original and adversarial TC trajectories at time $t$. For each location $(p,q)$, define its geodesic distance to the nearest location in $\mathcal{S}_t$ as
$ d_t(p,q)=\min_{(i,j)\in\mathcal{S}_t}\arccos\!\big(\sin\phi_p\sin\phi_i+\cos\phi_p\cos\phi_i\cos(\lambda_q-\lambda_j)\big)$ if $\mathcal{S}_t\neq\emptyset$, or 0 otherwise, where $(\phi_i,\lambda_j)$ is the corresponding latitude-longitude of $(i,j)$. The geodesic distance to nearest target is used to compute the following pair of distance-based weights for each location $(p,q)$:
\begin{equation}
    \begin{aligned}
    w^{\mathrm{grad}}_{tpq} &=
    \begin{cases}
    1, & (p,q)\in \mathcal{S}_t,\\[4pt]
    \exp\!\big(-\tfrac{d_t(p,q)^2}{2\sigma_{\mathrm{grad}}^2}\big), & (p,q)\notin \mathcal{S}_t,
    \end{cases}\\[6pt]
    w^{\mathrm{reg}}_{tpq} &=
    \begin{cases}
    0, & (p,q)\in \mathcal{S}_t,\\[4pt]
    1-\exp\!\big(-\tfrac{d_t(p,q)^2}{2\sigma_{\mathrm{reg}}^2}\big), & (p,q)\notin \mathcal{S}_t.
    \end{cases}
    \end{aligned}
\label{eq:weighting_scheme_for_adversarial_attack}
\end{equation}
We use $w^{\mathrm{grad}}$ to modify the gradient  update step of standard PGD as follows: 
\begin{equation}
    \hat{Y}_{tij}^{'(k+1)} \;=\;\operatorname{Clip}_{\delta}\!\left(\hat{Y}_{tij}^{'(k)} - \eta \cdot w_{tij}^{\mathrm{grad}} \cdot \operatorname{sign}\!\left(\frac{\partial \mathcal{L}_{\text{adv}}}{\partial \hat{Y}_{tij}^{'(k)}}\right)\right),
\label{eq:adv_attack_process}
\end{equation}
where $k$ is the iteration number, $\eta$ is the step size, $\operatorname{sign}(\cdot)$ denotes element-wise sign function and $\operatorname{Clip}_{\delta}(\cdot)$ projects the perturbed input onto the $\ell_\infty$-ball of radius $\delta$ around $\hat{Y}_{tij}$. In this scheme, the locations in $\mathcal{S}_t$ always receive full gradient updates ($w^{\mathrm{grad}}_{tpq}=1$) while other locations are updated with exponentially decaying weights based on their geodesic distance to $\mathcal{S}_t$.
Additionally, the loss function is modified to incorporate a distance-based regularization penalty as follows:
\begin{align}
    \mathcal{L}_{\text{adv}}=-\frac{1}{\beta}\sum_{t=1}^{\beta}\sum_{i=1}^r\sum_{j=1}^c &\big[(1-\hat{P}_{tij}')^{\gamma(M_{tij})}\hat{Z}_{tij}^{*D}\log(\hat{P}_{tij}') 
    \nonumber \\
    &+\hat{P}_{tij}'^{\gamma(M_{tij})}(1-\hat{Z}_{tij}^{*D})\log(1-\hat{P}_{tij}')\big] \nonumber\\
    &+\lambda \ \|w^{\mathrm{reg}}_{tij}(\hat{Y}_{tij}-\hat{Y}_{tij}')\|_2^2 ,
\label{eq:adv_objective_function}
\end{align}
where $\hat{P}_{tij}'$ is the surrogate's predicted probability for adversarial input $\hat{\mathbf{Y}}'$, $\hat{Z}_{tij}^{*D}$ is the dilated adversarial target, 
and $\lambda$ is the regularization hyperparameter. The distance-based regularization promotes stronger perturbations near the target region $\mathcal{S}_t$ and weaker perturbations at locations farther away.
The focal loss modulating parameter $\gamma$ depends on the calibration mask: $\gamma(M_{tij})=2$ if $M_{tij}=0$ and $\gamma(M_{tij})=0$ if $M_{tij}=1$. 
For locations with $M_{tij}=0$, since the classification by the surrogate model is correct,  we use the default focal-loss setting $\gamma=2$. In contrast, $M_{tij}=1$ marks locations misclassified by the surrogate model. These locations could mislead the attack algorithm into believing the adversarial objective has already been satisfied. Since the original predictions at these locations are incorrect, 
using a large $\gamma$ would suppress their gradients. To prevent this and ensure the optimization algorithm focuses on correcting these potentially misclassified locations, we set $\gamma=0$.

\section{End-to-end Upstream Attack: Generating Adversarial Input Samples}
\label{sec:upstream_attack}

Unlike the downstream attack method, which directly modifies the upstream forecast, the upstream attack method aims to construct an adversarial input $\mathbf{X}'$ to the DLWF model so that its output forecast $f(\mathbf{X}')$, when processed by the downstream TC detection software $g$, produces a trajectory $g(f(\mathbf{X}'))$ that aligns closely with the adversarial target $\hat{Z}^{*}$. However, constructing the adversarial input is challenging because the perturbations are introduced only over a short input window of length $\tau$, yet they must propagate through the DLWF model and affect the forecast trajectory over a substantially longer forecast horizon $\beta$, with $\beta \gg \tau$. 

To address this challenge, we propose a multi-stage optimization workflow for the end-to-end upstream attack. The first stage applies {\textit{Cyc-Attack}} to generate an adversarial upstream forecast $\hat{Y}'$. The second stage leverages $\hat{\mathbf{Y}}'$ as a reference to construct an initial adversarial upstream input $\mathbf{X}'$.
Finally, the third stage refines $\mathbf{X}'$ through end-to-end optimization, producing an adversarial input whose forecast leads to a TC trajectory aligned with the desired adversarial target.

During the second stage, we optimize the adversarial input $\mathbf{X}'$ by minimizing the mean squared error between its resulting upstream forecast $f(\mathbf{X}')$ and adversarial upstream forecast $\hat{\mathbf{Y}}'$ over regions associated with the original and target TC trajectories, while limiting perturbations in regions unrelated to the attack objective. Specifically, the objective function for stage 2 is defined as:
\begin{equation}
\mathcal{L}_{\text{stage-2}}=\mathcal{L}_{\mathrm{guide}}+\lambda_{\mathrm{unrelated}}\mathcal{L}_{\mathrm{unrelated}}+\lambda_{\mathrm{reg}}\mathcal{L}_{\mathrm{reg}},
\label{eq:upstream_guidance_objective}
\end{equation}
where $\mathcal{L}_{\mathrm{guide}}=(\mathcal{L}_{\mathrm{tgt}}+w_o\mathcal{L}_{\mathrm{orig}})/(1+w_o)$ is a weighted combination of the losses for the regions associated with the target and original trajectories. The corresponding loss for each region $r\in\{\mathrm{tgt},\mathrm{orig}\}$ is
\begin{equation}
\mathcal{L}_{r}=\frac{1}{\beta}\sum_{t=1}^{\beta}\frac{\sum_{i,j}w_{r,tij}\|f(X')_{tij}-\hat{Y}'_{tij}\|^{2}}{\sum_{i,j}w_{r,tij}},
\label{eq:upstream_region_guidance}
\end{equation}
where 
$\beta$ is the forecast horizon and $w_{r,tij}$ combines a spatial mask with Gaussian distance-based weights to emphasize locations near the corresponding trajectory. The original region excludes locations overlapping with the target region. The regularization term $\mathcal{L}_{\mathrm{reg}}$ is analogous to the third term in Equation~\eqref{eq:adv_objective_function}, applied to the input perturbation, while $\mathcal{L}_{\mathrm{unrelated}}$ penalizes deviations of $f(\mathbf{X}')$ from the original forecast $f(\mathbf{X})$ outside the manipulated TC regions. We optimize $\mathbf{X}'$ using PGD with the distance-based gradient weighting strategy introduced in Equation~\eqref{eq:adv_attack_process}, adapted to the upstream input space and augmented with variable-specific Gaussian gradient smoothing. The perturbation that minimizes $\mathcal{L}_{\text{stage-2}}$ is used to initialize the third stage.

In the third stage, we further refine the adversarial upstream input initialized in the second stage with an end-to-end optimization to find an adversarial input that will produce TC trajectories more closely aligned with the adversarial target $\hat{Z}^{*}$. The objective function for stage 3 is given by
\begin{equation}
\mathcal{L}_{\text{stage-3}}=\mathcal{L}_{\mathrm{traj}}+\lambda_{\mathrm{reg}}\mathcal{L}_{\mathrm{reg}},
\label{eq:upstream_trajectory_objective}
\end{equation}
where $\mathcal{L}_{\mathrm{traj}}$ corresponds to the calibrated focal loss introduced in the first two terms in Equation~\ref{eq:adv_objective_function}, while $\mathcal{L}_{\mathrm{reg}}$ is defined as in the second stage. We employ the same gradient-smoothed, distance-weighted PGD strategy to minimize $\mathcal{L}_{\text{stage-3}}$.

\section{Experimental Evaluation}
\label{sec:Experimental results}

\subsection{Experimental Setup}
\label{sec:Experimental Setup}

\subsubsection{Datasets}

The datasets used in our experiments are obtained from the following two data sources: \textit{\textbf{IBTrACS}} (International Best Track Archive for Climate Stewardship), which provides historical TC track locations at 6-hourly intervals, and \textit{\textbf{ERA5 Reanalysis Data}}, which provides hourly global weather fields 
at $1^{\circ}\times1^{\circ}$ resolution. 
The ERA5 reanalysis data
include the following variables: 
(1) {five} single-level variables — \textit{2-meter temperature} (K), \textit{mean sea level pressure} (Pa), \textit{10-meter u-component of wind} (m/s), \textit{10-meter v-component of wind} (m/s), and \textit{total precipitation} (m);  
(2) {six} multi-level variables — \textit{temperature} (K), \textit{geopotential} (m\textsuperscript{2}/s\textsuperscript{2}), \textit{u-component of wind} (m/s), \textit{v-component of wind} (m/s), \textit{vertical velocity} (Pa/s), and \textit{specific humidity} (kg/kg), provided at {thirteen} pressure levels (50, 100, 150, 200, 250, 300, 400, 500, 600, 700, 850, 925, and 1000 hPa);  
(3) {three} forcing variables — \textit{total-of-atmosphere incident solar radiation} (J/m\textsuperscript{2}), \textit{land-sea mask} (unitless), and \textit{surface geopotential} (m\textsuperscript{2}/s\textsuperscript{2});  
and (4) {four} post-calculated variables — \textit{sine and cosine of year progress}, and \textit{sine and cosine of day progress}, which are derived from timestamps and are unitless. 

Using these sources, we constructed two datasets, \textit{TC1} and \textit{TC2}. \textit{TC1} consists of 285 tropical cyclones (TCs) recorded in IBTrACS between January 1, 2019, and January 1, 2025. For each TC, the ERA5 fields from the first two 6-hourly time steps, beginning at the time of TC formation, are used as input to the \textit{GraphCast} model. \textit{GraphCast} then generates forecasts for the subsequent 12 time steps (72 hours). These forecasts are subsequently processed by \textit{TempestExtremes} to extract the corresponding downstream TC trajectory. \textit{TC2} contains the trajectories for 10 major TCs selected for our case studies, using the same procedure as for \textit{TC1}, except that the forecast window, $\beta$, for \textit{GraphCast} and \textit{TempestExtremes} is extended from 12 time steps (72 hours) to 38 time steps (228 hours). 
The predicted TC trajectories are used to construct adversarial downstream targets. %
For surrogate model training, the \textit{TC1} samples were split into training, validation, and test sets using a $0.6{:}0.2{:}0.2$ ratio. After excluding maps without TC events, the resulting training, validation, and test sets contained $2052$, $684$, and $684$ TC location maps, respectively.

\subsubsection{Adversarial Target Trajectory Construction} 

We construct adversarial TC trajectories by progressively steering the original predicted tracks toward alternative directions while preserving their initial locations, temporal resolution, and geodesic step lengths. Specifically, at each time step, we select a direction from eight compass directions based on its deviation from the original movement and its consistency with the preceding adversarial movement. The selected direction is blended with the original direction to encourage smooth trajectory changes, and the next location is computed using great-circle projection.  For samples containing multiple TCs, we construct adversarial targets by modifying one random trajectory while keeping all others unchanged. Technical details on adversarial target construction are provided in Supplementary Materials C.

\subsubsection{Baseline Methods} 

For downstream attack, we compare the performance of {\textit{Cyc-Attack}} against four baselines: \textbf{\textit{RGE}}~\cite{nesterov2017random}, a gradient-free method based on zeroth-order optimization to estimate gradients via randomized queries; \textbf{\textit{ALA}}~\cite{ruan2023vulnerability}, which uses an Adam-based optimizer; \textbf{\textit{TAAOWPF}}~\cite{heinrich2024targeted}, which uses PGD-based updates; and \textbf{\textit{AOWF}}~\cite{imgrund2025adversarial}, which applies an Adam-based optimizer with cosine-scaled step-size schedule. All methods use \textit{DeepLabV3+}~\cite{chen2018encoder} as the surrogate model  for attacking the \textit{TempestExtremes} TC detector. Each attack is run for a maximum of 1{,}000 iterations, with early stopping triggered if the loss does not change significantly for 15 consecutive iterations. We set the clipping threshold to $\delta_{\text{downstream}}=10.0$ and the step size to $\eta=0.01$. Additional training and tuning details are described in Supplementary Materials D. {A larger clipping threshold is required because of the chaotic dynamics of the atmosphere. As the initial perturbations can amplify over time, the long forecasting horizon (38 lead time steps) makes it more difficult to maintain sufficient perturbation strength to influence the downstream TC detector. A larger perturbation budget therefore provides the attack with enough flexibility to remain effective throughout the forecast trajectory.}


For upstream attacks, we compare \textit{Cyc-Attack} against ALA, the best-performing baseline in downstream attacks, as will be demonstrated later in our experiments. The attacked variables include mean sea level pressure; zonal and meridional wind components at 10 m and at the 300, 500, 700, and 850 hPa pressure levels; along with temperature and geopotential heights at 300 and 500 hPa pressure levels. We allow a maximum of 1,000 iterations for ALA and for each of the two phases of \textit{Cyc-Attack}. The perturbation clipping threshold is set to $\delta_{\mathrm{upstream}}=3.0$ in standardized input space. 

\subsubsection{Evaluation Metrics} 

We evaluate the competing methods using three criteria. 
(1) \textbf{\emph{Location-level accuracy}}, which compares the adversarial target locations against the predicted TC locations in terms of their true positive rates (TPR), true negative rates (TNR), false positive rates (FPR), and false negative rates (FNR). For each hurricane, we aggregate the set of forecasted trajectory maps, $g(\hat{\mathbf{Y}}')$, into a single binary map, $\hat{\mathbf{Z}}' \in \{0,1\}^{r \times c}$, where the $(i,j)$-th entry is set to 1 if a TC event is predicted at that location at any forecast time step, and 0 otherwise. The evaluation metrics are then computed by comparing these time-aggregated maps with the corresponding adversarial target maps.
(2) \textbf{\emph{Trajectory-level accuracy}}, measured by TC trajectory detection rate (DR) and false alarm rate (FAR). Two locations are considered matched if their great-circle distance is less than $1^\circ$ for TC1 and $2^\circ$ for TC2. A target trajectory is considered successfully detected if at least 50\% of its locations overlap with the closest predicted trajectory. Conversely, a predicted trajectory is classified as a false alarm if it has no overlap with any target trajectory or overlaps only during the initial time steps before subsequently diverging. Note that, unlike location-level accuracy, which is computed by aggregating the predicted TC locations across all forecast time steps into a single binary map, trajectory-level accuracy is computed by comparing the predicted and target locations at each forecast time step.
(3) \textbf{\emph{Closeness}} ($\delta_C$), which is computed based on the $\ell_1$-norm between the original upstream forecast $\hat{\mathbf{Y}}$ and its adversarial counterpart $\hat{\mathbf{Y}}'$. A smaller $\delta_C$ indicates a stealthier attack.

\begin{table}[bt]
\centering
\small
\caption{Impact of varying the dilation radius \(R_S\) on the performance of a pre-trained \textit{DeepLabV3+} (Xception backbone) surrogate model. The evaluation is conducted on a test set of 684 global 1$^\circ$ maps (\(180 \times 360\) grid each).
Note that $R_S = 0$ corresponds to applying standard focal loss (no dilation).}
\label{tab:diff_R_surrogate_rates}
\begin{tabular}{ccccc}
\toprule
$R_S$ & TNR$\uparrow$ & FNR$\downarrow$ & FPR$\downarrow$ & TPR$\uparrow$ \\
\midrule
0 & 0.9933 & 0.0104 & 0.0067 & 0.9896 \\
1 & 0.9999 & 0.4135 & 0.0001 & 0.5865 \\
2 & 0.9998 & 0.1869 & 0.0002 & 0.8131 \\
3 & 0.9993 & 0.0692 & 0.0007 & 0.9308 \\
5 & 0.9986 & 0.0606 & 0.0014 & 0.9394 \\
\bottomrule
\end{tabular}
\end{table}

\begin{figure}[bt]
\centering
\includegraphics[width=1.0\linewidth]{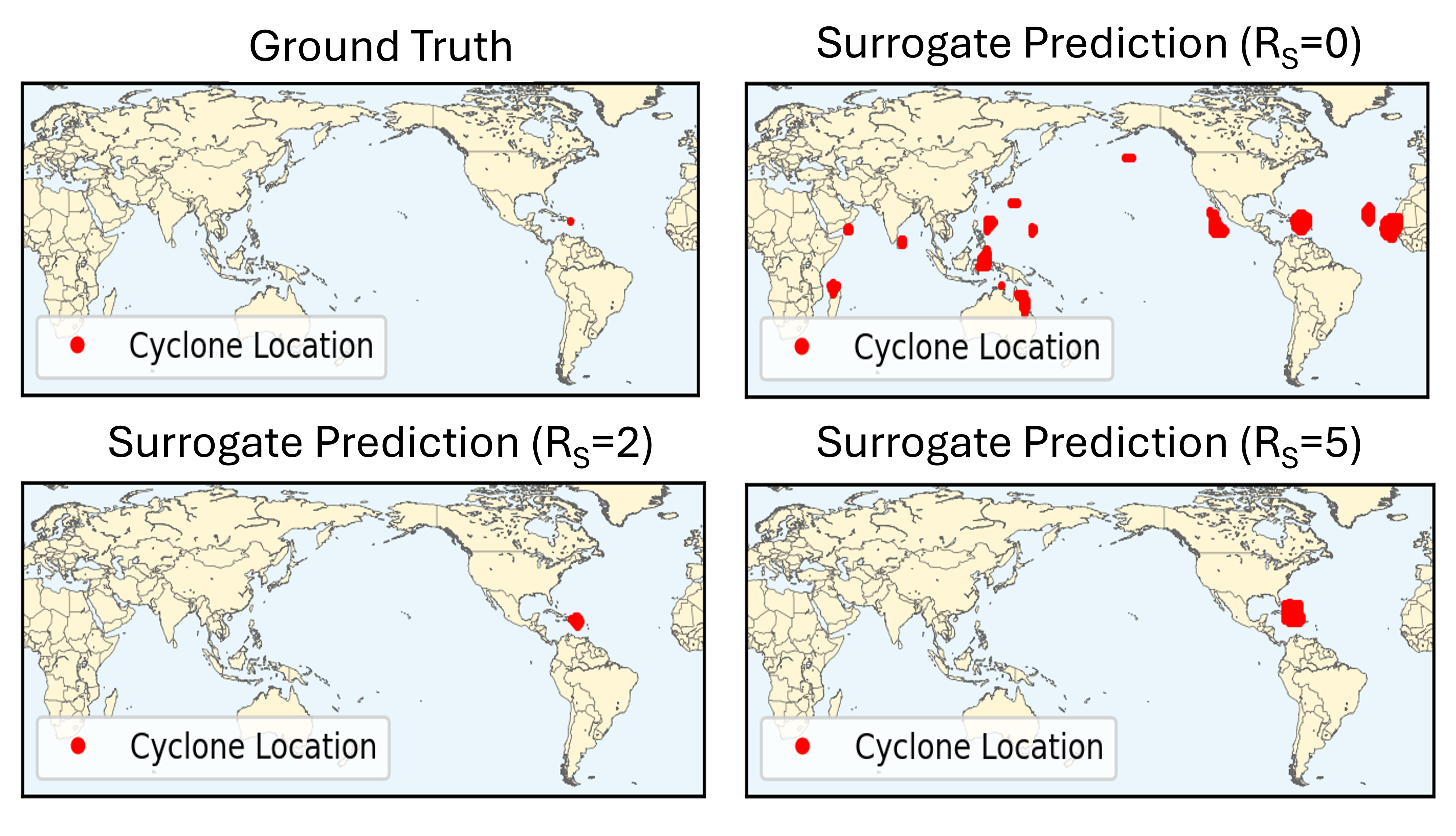}
\caption{Visualization of TC location predictions using \textit{DeepLabV3+} (\textit{Xception} backbone) trained on labels produced with different dilation radii. \(R_S=0\) corresponds to standard focal loss (no dilation). Predictions are compared against ground-truth detections from \textit{TempestExtremes} (top-left). 
}
\label{fig:surrogate_different_radius}
\end{figure}

\subsection{Experimental Results}
\label{sec:Experimental Results}

\subsubsection{Surrogate Model Performance}

Table~\ref{tab:diff_R_surrogate_rates} summarizes the accuracy of the surrogate model in reproducing the outputs of \textit{TempestExtremes} under different dilation radii $R_S$. Without dilation ($R_S=0$), the model achieves the highest TPR but also generates numerous false alarms scattered across the globe, as shown in Figure~\ref{fig:surrogate_different_radius}. Increasing $R_S$ from $0$ (no dilation) to $1$ reduces spatially scattered false positives by concentrating predictions into compact clusters around TC locations. However, a further increase in $R_S$ may alter the trade-off between TPR and FPR. We choose $R_S=2$ for our pre-trained surrogate model, as it provides a balanced trade-off between true positive and false positive rates.

\begin{table*}[tb]
\centering
\footnotesize
\caption{Downstream attack performance of different methods 
in terms of their accuracies at location-level (FPR, FNR, TPR), trajectory-level (DR and FAR), and closeness ($\delta_C$). {\textit{Cyc-Attack}}-\(d\) and -\(w\) denote variants of {\textit{Cyc-Attack}} that omit the kernel-dilation strategy (Equation~\ref{eq:surrogate_dilation}) and the weighting scheme (Equation~\ref{eq:weighting_scheme_for_adversarial_attack}), respectively. 
For all reported {\textit{Cyc-Attack}} results, the dilation radius is {\(R_A=1\)}. For \textit{RGE}, results are available for TC2 only, as the method is computationally prohibitive for the massive TC1 dataset. The location-level and trajectory-level metrics are computed by pooling counts across all evaluation cases (micro-averaging), whereas the closeness metric is averaged across cases.}
\label{tab:tc1_tc2_results}
\begin{tabular}{l|ccc|cc|c}
\hline
Method & \multicolumn{3}{c|}{TC1 Location-level} & \multicolumn{2}{c|}{TC1 Trajectory-level} & TC1 Closeness \\ \cline{2-7}
 & FPR$\downarrow$ & FNR$\downarrow$ & TPR$\uparrow$ & DR$\uparrow$ & FAR$\downarrow$ & $\delta_C\downarrow$ \\
\hline
ALA
 & \textbf{0.0001}~($\pm$0.0000)
 & 0.8218~($\pm$0.0293)
 & 0.1782~($\pm$0.0293)
 & 0.3197~($\pm$0.0864)
 & 0.1393~($\pm$0.0572)
 & 0.0459~($\pm$0.0002) \\
TAAOWPF
 & 0.0002~($\pm$0.0000)
 & 0.8640~($\pm$0.0263)
 & 0.1360~($\pm$0.0263)
 & 0.1038~($\pm$0.0414)
 & 0.5991~($\pm$0.0610)
 & 0.1753~($\pm$0.0010) \\
AOWF
 & \textbf{0.0001}~($\pm$0.0000)
 & 0.7971~($\pm$0.0322)
 & 0.2029~($\pm$0.0322)
 & 0.2977~($\pm$0.0749)
 & 0.1527~($\pm$0.0585)
 & 0.0450~($\pm$0.0007) \\
\hline
Cyc-Attack
 & \textbf{0.0001}~($\pm$0.0000)
 & \textbf{0.7001}~($\pm$0.0352)
 & \textbf{0.2999}~($\pm$0.0352)
 & \textbf{0.5570}~($\pm$0.0793)
 & \textbf{0.0537}~($\pm$0.0345)
 & \textbf{0.0004}~($\pm$0.0000) \\
Cyc-Attack-\textit{d}
 & \textbf{0.0001}~($\pm$0.0000)
 & 0.8476~($\pm$0.0252)
 & 0.1524~($\pm$0.0252)
 & 0.2804~($\pm$0.0841)
 & 0.0935~($\pm$0.0573)
 & \textbf{0.0003}~($\pm$0.0000) \\
Cyc-Attack-\textit{w}
 & 0.0002~($\pm$0.0000)
 & 0.8421~($\pm$0.0253)
 & 0.1579~($\pm$0.0253)
 & 0.0655~($\pm$0.0324)
 & 0.5284~($\pm$0.0509)
 & 0.1850~($\pm$0.0010) \\
\hline \hline
Method & \multicolumn{3}{c|}{TC2 Location-level} & \multicolumn{2}{c|}{TC2 Trajectory-level} & TC2 Closeness \\ \cline{2-7}
 & FPR$\downarrow$ & FNR$\downarrow$ & TPR$\uparrow$ & DR$\uparrow$ & FAR$\downarrow$ & $\delta_C\downarrow$ \\
\hline
RGE
 & \textcolor{gray}{0.0003}
 & \textcolor{gray}{0.8994}
 & \textcolor{gray}{0.1006}
 & \textcolor{gray}{0}
 & \textcolor{gray}{1}
 & \textcolor{gray}{0.0277} \\
\hline
ALA
 & 0.0007~($\pm$0.0003)
 & 0.7380~($\pm$0.1061)
 & 0.2620~($\pm$0.1061)
 & 0.2222~($\pm$0.2083)
 & 0.5926~($\pm$0.2139)
 & 0.0467~($\pm$0.0005) \\
TAAOWPF
 & 0.0045~($\pm$0.0009)
 & 0.7812~($\pm$0.0725)
 & 0.2188~($\pm$0.0725)
 & 0.0211~($\pm$0.0190)
 & 0.8803~($\pm$0.0810)
 & 0.1949~($\pm$0.0049) \\
AOWF
 & 0.0012~($\pm$0.0005)
 & 0.7067~($\pm$0.1173)
 & 0.2933~($\pm$0.1173)
 & 0.1000~($\pm$0.1000)
 & 0.7000~($\pm$0.0876)
 & 0.0500~($\pm$0.0016) \\
\hline
Cyc-Attack
 & \textbf{0.0004}~($\pm$0.0002)
 & \textbf{0.4447}~($\pm$0.0594)
 & \textbf{0.5553}~($\pm$0.0594)
 & \textbf{0.7500}~($\pm$0.2056)
 & \textbf{0.0833}~($\pm$0.0926)
 & 0.0003~($\pm$0.0000) \\
Cyc-Attack-\textit{d}
 & \textbf{0.0004}~($\pm$0.0002)
 & 0.5962~($\pm$0.1119)
 & 0.4038~($\pm$0.1119)
 & 0.5000~($\pm$0.1594)
 & 0.2083~($\pm$0.1086)
 & \textbf{0.0002}~($\pm$0.0000) \\
Cyc-Attack-\textit{w}
 & 0.0056~($\pm$0.0007)
 & 0.6971~($\pm$0.0911)
 & 0.3029~($\pm$0.0911)
 & 0.0275~($\pm$0.0255)
 & 0.8901~($\pm$0.0525)
 & 0.2169~($\pm$0.0043) \\
\hline
\end{tabular}%
\end{table*}

\begin{figure*}[tb]
    \centering
    \includegraphics[width=1.0\textwidth]{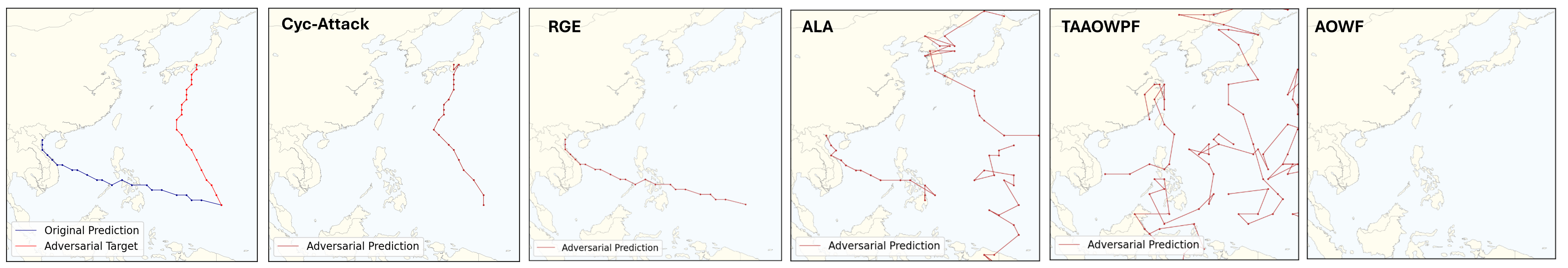}  
    \caption{Downstream attack visualization of {\textit{Cyc-Attack}} and baseline methods for typhoon \textit{Haiyan} (2013). The left panel shows the original \textit{TempestExtremes} trajectory from \textit{GraphCast} forecasts (\textcolor{blue}{blue}) and the adversarial target (\textcolor{red}{red}); the right panels show trajectories produced by \textit{Cyc-Attack} and baseline methods.} 
\label{fig:comparison_of_different_methods}
\end{figure*}

\subsubsection{Downstream Attack Performance Comparison}

Table~\ref{tab:tc1_tc2_results} compares {\textit{Cyc-Attack}} against the baseline methods in terms of their predicted TC trajectory accuracy and closeness to the original upstream forecast. 
Each metric is accompanied by a nonparametric bootstrap $95\%$ confidence interval computed from 10,000 paired resamples~\cite{koehn2004statistical,efron1992bootstrap}. The zeroth-order baseline \textit{RGE} is ineffective\footnote{For the RGE method, attacking the 10 samples in \textit{TC2} already takes approximately 1.5 hours on a single Google T4 GPU with 22.5 GB memory and still shows no clear convergence. In contrast, other gradient-based methods require about 3 seconds per iteration on average with clear convergence.} in this setting, with DR equal to $0$ and FAR equal to $1$; therefore, we exclude it from the main comparison. Among the remaining methods, {\textit{Cyc-Attack}} achieves the best overall performance, attaining the highest location-level accuracy and trajectory-level DR, while achieving the lowest FAR and $\delta_C$. In contrast, the baselines (\textit{AOWF, ALA, TAAOWPF}) either fail to remove the original trajectory or cannot reliably generate the adversarial target, as demonstrated in Figure~\ref{fig:comparison_of_different_methods} for typhoon Haiyan. Results for other hurricanes in \textit{TC2} are provided in Supplementary Materials E. 
To evaluate the statistical significance of the performance improvements, we compare \textit{Cyc-Attack} against the strongest non-ablation baseline, ALA, using two-sided paired $t$-tests, with results presented in Table~\ref{tab:tc1_tc2_results_sig}. On \textit{TC1}, the improvements achieved by \textit{\textit{Cyc-Attack}} are statistically significant $(p < 0.05)$ for all metrics except FPR and FAR. On \textit{TC2}, the improvements are statistically significant for all 6 metrics. 

\subsubsection{Sensitivity Analysis and Ablation Studies} 

The performance improvements observed in \textit{Cyc-Attack} are primarily attributed to the use of kernel dilation and distance weighting to localize the perturbations around the adversarial target, suppress false-positive trajectories, and improve spatial coherence. Without these strategies, existing gradient-based attacks would produce scattered perturbations or poor target alignment, as shown in Figure~\ref{fig:comparison_of_different_methods}. To quantify the contribution of each strategy, we consider two variations of \textit{Cyc-Attack}. Specifically, \textit{Cyc-Attack}-(d) disables the kernel-dilation strategy (Equation~\eqref{eq:surrogate_dilation}), while \textit{Cyc-Attack}-(w) disables the distance-weighting scheme (Equation~\eqref{eq:weighting_scheme_for_adversarial_attack}). All other components are kept identical to those of the full \textit{Cyc-Attack} method. As shown in Table~\ref{tab:tc1_tc2_results}, both ablation variants achieve lower DR and higher FAR than the full \textit{Cyc-Attack} method, while removing distance weighting also results in substantially larger perturbations. These results demonstrate that both kernel dilation and distance weighting strategies play a significant role to improve downstream attack performance.

\begin{table*}[tb]
\centering
\footnotesize
\setlength{\tabcolsep}{4.5pt}
\renewcommand{\arraystretch}{0.95}
\caption{Statistical significance of performance improvement for {\textit{Cyc-Attack}} over \textit{ALA} using two-sided paired $t$-tests. Unlike Table~\ref{tab:tc1_tc2_results}, which reports pooled metrics, this table reports macro-averaged per-case metrics over matched evaluation cases used in the paired t-tests.}
\label{tab:tc1_tc2_results_sig}
\begin{tabular}{lcccccccc}
\toprule
& \multicolumn{4}{c}{Dataset TC1} & \multicolumn{4}{c}{Dataset TC2} \\
\cmidrule(lr){2-5}\cmidrule(lr){6-9}
Metric
& Cyc-Attack & ALA & $t$ & $p$
& Cyc-Attack & ALA & $t$ & $p$ \\
\midrule
FPR$\downarrow$
& \textbf{0.0001} & \textbf{0.0001} & 0.376 & 0.7078
& \textbf{0.0004} & 0.0007 & $-2.708$ & 0.0241 \\
FNR$\downarrow$
& \textbf{0.6597} & 0.8292 & $-5.915$ & $2.21{\times}10^{-8}$
& \textbf{0.3963} & 0.8275 & $-4.677$ & 0.0012 \\
TPR$\uparrow$
& \textbf{0.3403} & 0.1708 & 5.915 & $2.21{\times}10^{-8}$
& \textbf{0.6037} & 0.1725 & 4.677 & 0.0012 \\
DR$\uparrow$
& \textbf{0.5875} & 0.3167 & 3.638 & 0.0005
& \textbf{0.7250} & 0.1375 & 4.222 & 0.0039 \\
FAR$\downarrow$
& \textbf{0.0333} & 0.0917 & $-1.712$ & 0.0909
& \textbf{0.0875} & 0.7250 & $-5.493$ & 0.0009 \\
$\delta_C$$\downarrow$
& \textbf{0.0004} & 0.0459 & $-195.303$ & $1.85{\times}10^{-180}$
& \textbf{0.0003} & 0.0467 & $-89.013$ & $1.44{\times}10^{-14}$ \\
\bottomrule
\end{tabular}
\end{table*}

\begin{figure}[tb]
\centering
\includegraphics[width=\linewidth]{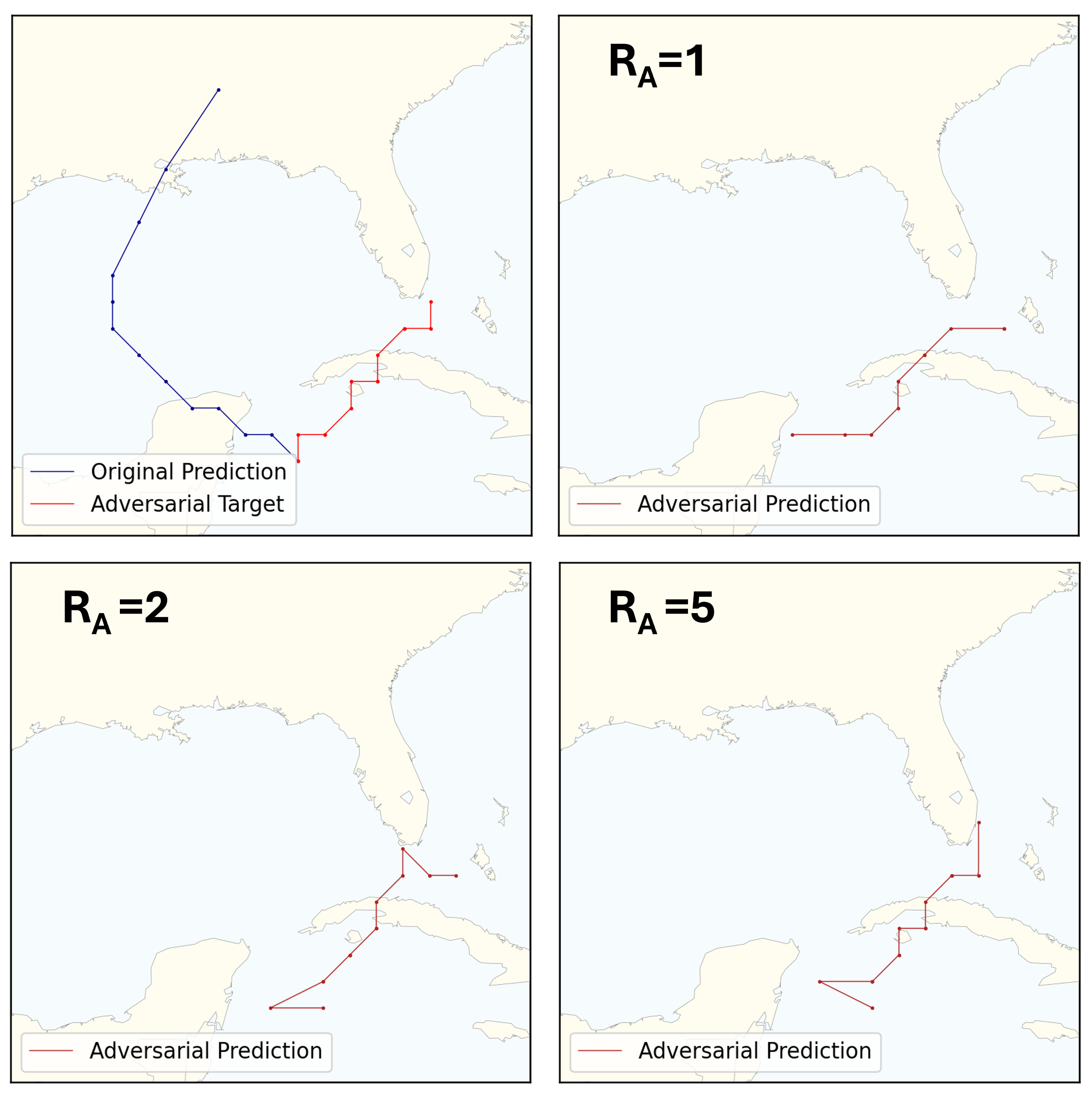}
\caption{Effect of varying the kernel dilation radius $R_A$ on the adversarial trajectories generated by the {\textit{Cyc-Attack}} method for downstream attack on hurricane \textit{Zeta} (2020).}
\label{fig:adv_attack_different_radius}
\end{figure}

We further analyze the sensitivity of {\textit{Cyc-Attack}} to the dilation radius $R_A$. The results for hurricane \textit{Zeta}, shown in Figure~\ref{fig:adv_attack_different_radius}, suggest that larger values of $R_A$ tend to produce more zigzag trajectories.  This occurs because a larger dilation region allows perturbations to extend beyond the target trajectory, creating additional candidate locations for TC events. As the number of candidate TC locations increases, the generated trajectories become more prone to exhibiting zigzag patterns. Additional results can be found in Supplementary Materials F. 

\subsubsection{Upstream Attack Performance Comparison} 

\begin{figure}[tb]
\centering
\includegraphics[width=1.0\linewidth]{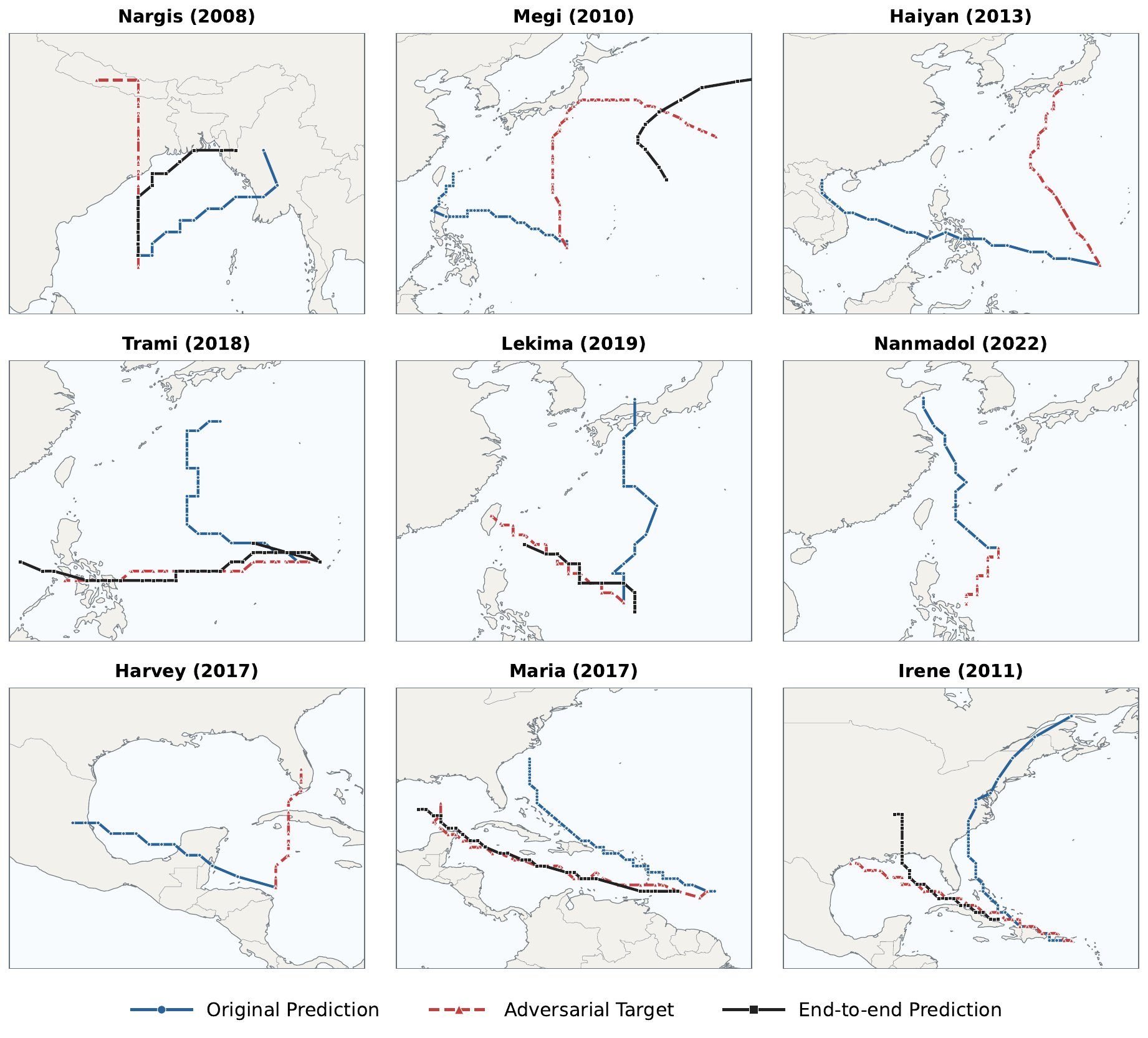}
\caption{{Upstream attack performance of \textit{Cyc-Attack} on TC2 with adversarial targets at a mean distance of 1146.5054 km from the original trajectories.}
}
\label{fig:end2end_cyc_attack_hard}
\end{figure}

\begin{figure}[tb]
\centering
\includegraphics[width=1.0\linewidth]{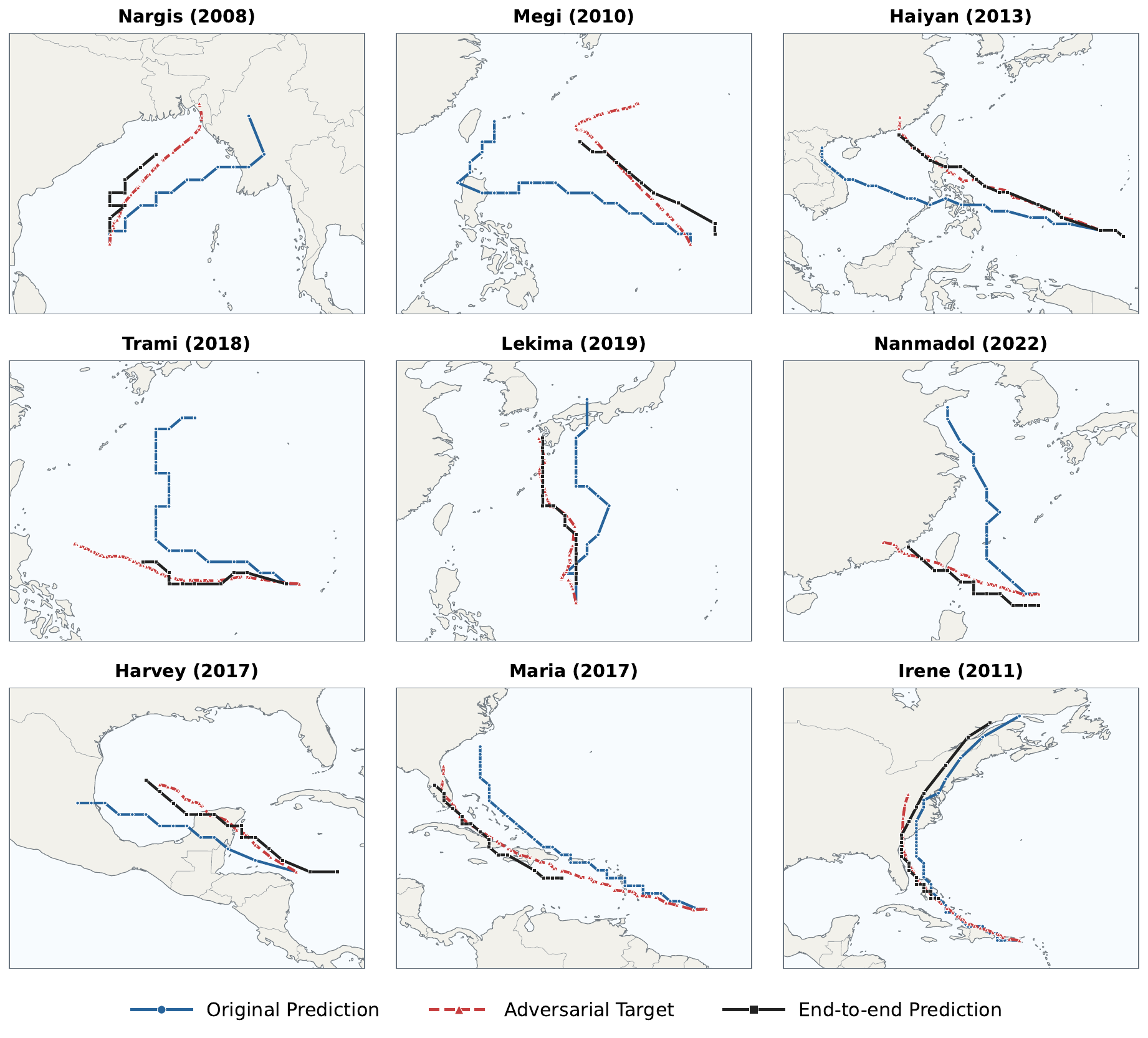}
\caption{{Upstream attack performance of \textit{Cyc-Attack} on TC2 with adversarial targets at a mean distance of 439.8060 km from the original trajectories.}
}
\label{fig:end2end_cyc_attack_easier}
\end{figure}

Table~\ref{tab:cyc_attack_ala_comparison} compares the upstream attack performance of our end-to-end \textit{Cyc-Attack} framework against the end-to-end ALA method on the \textit{TC2} dataset. For reference, we also report the results for their corresponding downstream models. We first observe that the end-to-end models generally achieve weaker performance than their downstream counterparts. This is not surprising, as their adversarial upstream forecasts $\hat{\mathbf{Y}}'$ are not constrained by the DLWF model, allowing greater flexibility in perturbing the data without accounting for whether the resulting forecasts can be realistically generated by GraphCast. Nevertheless, a comparison between the two end-to-end approaches shows that \textit{Cyc-Attack}-e2e consistently outperforms ALA-e2e across all evaluated metrics. Notably, ALA-e2e achieves DR of only $0.040$, which is substantially lower than that of \textit{Cyc-Attack}-e2e ($0.6471$). 
Visualizations of the adversarial attack trajectories obtained by the end-to-end \textit{Cyc-Attack} for tropical cyclones in TC2 data are shown in Figures~\ref{fig:end2end_cyc_attack_hard} and \ref{fig:end2end_cyc_attack_easier}.


\begin{table}[tb]
\centering
\caption{Upstream attack performance comparison on the TC2 dataset. The suffix \texttt{-e2e} denotes an end-to-end attack model, while \texttt{-ds} indicates the downstream model only. $\tilde{\delta}_C$, applied to $\mathbf{X}$, is unavailable for downstream models since they do not produce adversarial input samples.}
\label{tab:cyc_attack_ala_comparison}
\footnotesize
\setlength{\tabcolsep}{5pt}
\begin{tabular}{lrrrrrr}
\toprule
Method & FPR$\downarrow$ & FNR$\downarrow$ & TPR$\uparrow$ & DR$\uparrow$ & FAR$\downarrow$ & $\tilde{\delta}_C$$\downarrow$ \\
\midrule
Cyc-At-e2e & 0.0002 & 0.6202 & 0.3798 & 0.6471 & 0.2353 & 0.0502 \\
ALA-e2e & 0.0003 & 0.9880 & 0.0120 & 0.0400 & 0.6400 & 0.4337 \\
Cyc-At-ds & 0.0004 & 0.4447 & 0.5553 & 0.7500 & 0.0833 & N/A \\
ALA-ds & 0.0007 & 0.7380 & 0.2620 & 0.2222 & 0.5926 & N/A\\
\bottomrule
\end{tabular}
\end{table}

\begin{figure}[tb]
\centering
\includegraphics[width=0.98\linewidth]{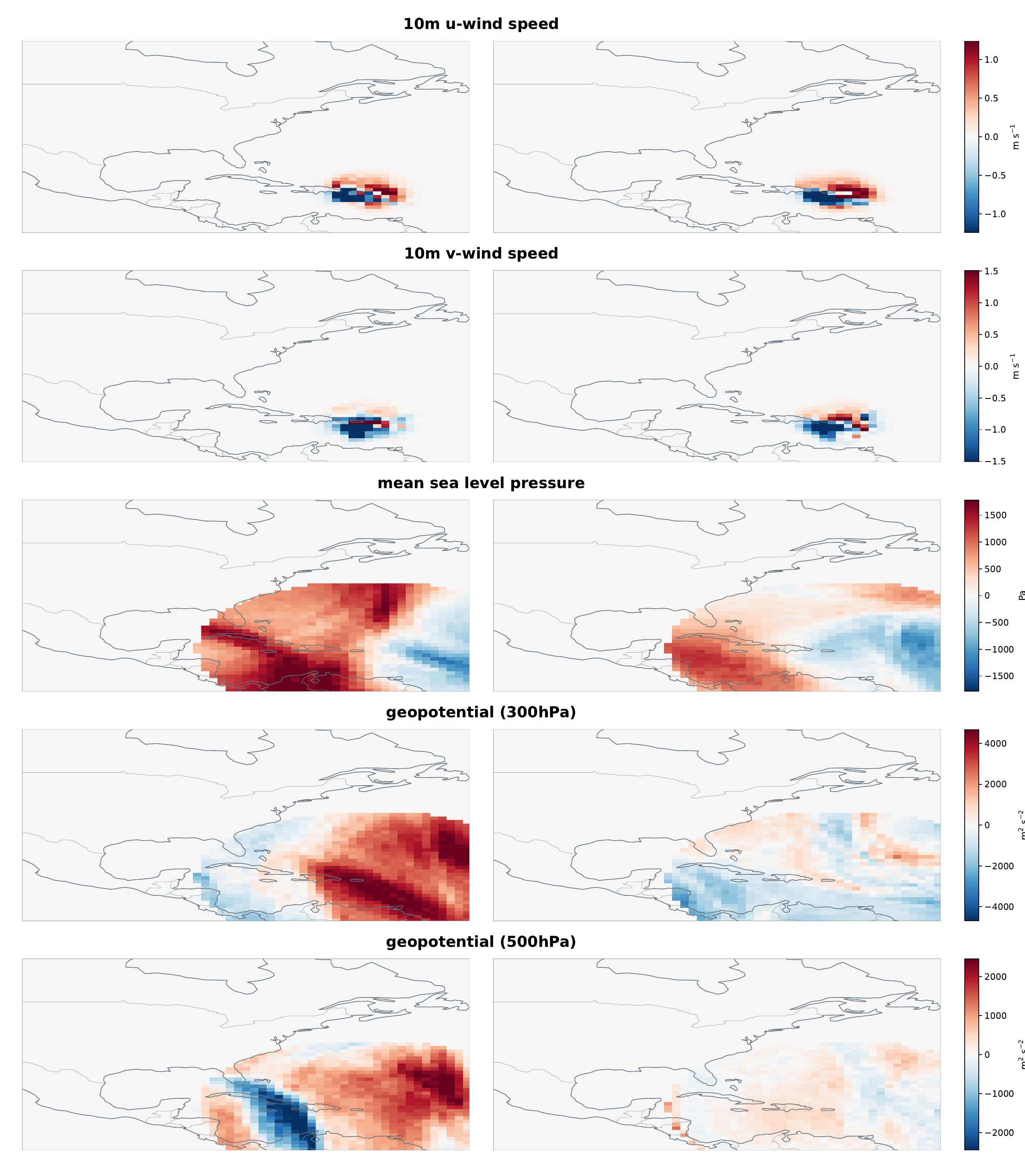}
\caption{Perturbations to the \textit{GraphCast} input by upstream attack on hurricane \textit{Irene} (2011). The two columns show the differences between the adversarial upstream input generated by the upstream attack and the original \textit{GraphCast} input at the two input time steps, respectively. Each row corresponds to a meteorological variable perturbed by the upstream attack and related to the downstream weather fields used by \textit{TempestExtremes}. All perturbations are presented in their original physical units.}
\label{fig:irene_physical_plausibility}
\end{figure}

We further analyze the perturbations generated by \textit{Cyc-Attack} across the two input windows of GraphCast using hurricane \textit{Irene} as case study. Figure~\ref{fig:irene_physical_plausibility} shows the perturbations applied to the wind speed, mean sea level pressure (MSLP), and geopotential height variables to achieve the targeted trajectory of hurricane \textit{Irene} (2011) shown in Figure~\ref{fig:end2end_cyc_attack_hard}. Note that the perturbation magnitudes for all variables remain within three standard deviations, as the clipping threshold is set to $3$. The wind components exhibit relatively larger localized perturbations, reflecting the need to induce transitions from non-extreme to extreme conditions associated with hurricane formation. In contrast, the perturbations to MSLP and geopotential heights are generally smaller in magnitude and spatially smoother. 

To evaluate the physical realisticness of the perturbed values, Table~\ref{tab:physical_range_comparison} summarizes the overall statistics of the original \textit{GraphCast} inputs and ground-truth forecasts in the TC1 dataset, focusing on five meteorological variables directly relevant to the downstream TC detection task and subject to upstream attacks: 10-m zonal and meridional winds, mean sea level pressure (MSLP), and geopotential at 300 and 500 hPa, and the statistics of adversarial input in the TC2 dataset. For each variable, we report the range of observed values across the 285 tropical cyclones in the TC1 dataset and compare it with the corresponding range of the adversarial inputs. 
The results suggest that the minimum and maximum adversarial (perturbed) values for wind speed and mean sea level pressure by \textit{Cyc-Attack} generally fall within the statistical range of the TC1 data. {The perturbed geopotential values slightly exceed this range. However, such perturbation could likely be mitigated by reducing the clipping threshold used in the upstream attack.} 


\begin{table}[t]
\centering
\caption{Physical ranges of meteorological variables in TC1 and corresponding adversarial values in TC2.}
\label{tab:physical_range_comparison}
\resizebox{\columnwidth}{!}{%
\begin{tabular}{lcc}
\toprule
{Variable} & {TC1 Ground-truth Range} & {TC2 Adversarial Range} \\
\midrule
10-m $|u|$ wind component (m/s) & 0--36.36 & 0--31.52 \\
10-m $|v|$ wind component (m/s) & 0--36.74 & 0--27.02 \\
Mean sea level pressure (Pa) & 91,303.00--106,822.00 & 93,162.31--105,149.50 \\
Geopotential @300hPa ($m^2s^{-2}$) & 75,787.81--96,787.50 & 76,271.75--104,379.87 \\
Geopotential @500hPa ($m^2s^{-2}$) & 43,307.90--59,109.62 & 43,784.96--63,338.88 \\
\bottomrule
\end{tabular}
}
\end{table}

\subsubsection{Adversarial Target Difficulty} 

{Despite its strong performance, the end-to-end \textit{Cyc-Attack} framework does not always succeed in generating the attacker's desired target trajectory, even when the downstream model is capable of reproducing that trajectory accurately. This limitation arises for two main reasons. First, the desired target trajectory may be located too far from the original track, requiring perturbations that exceed the available perturbation budget (i.e., the clipping threshold). In such cases, it may not be possible to construct an adversarial input that steers the forecast to the desired target while satisfying the perturbation constraint. Second, GraphCast generates a 38-step forecast from only two input time steps. Consequently, perturbing only these two input states may not provide sufficient control over the entire forecast trajectory. Although the downstream attack can identify an adversarial upstream forecast that produces the desired target trajectory, that adversarial forecast may not be realizable by GraphCast from any feasible perturbation of its two input time steps.}

\begin{figure}[tb]
\centering
\setlength{\fboxrule}{0.6pt}
\setlength{\fboxsep}{1pt}
\includegraphics[width=1.0\linewidth]{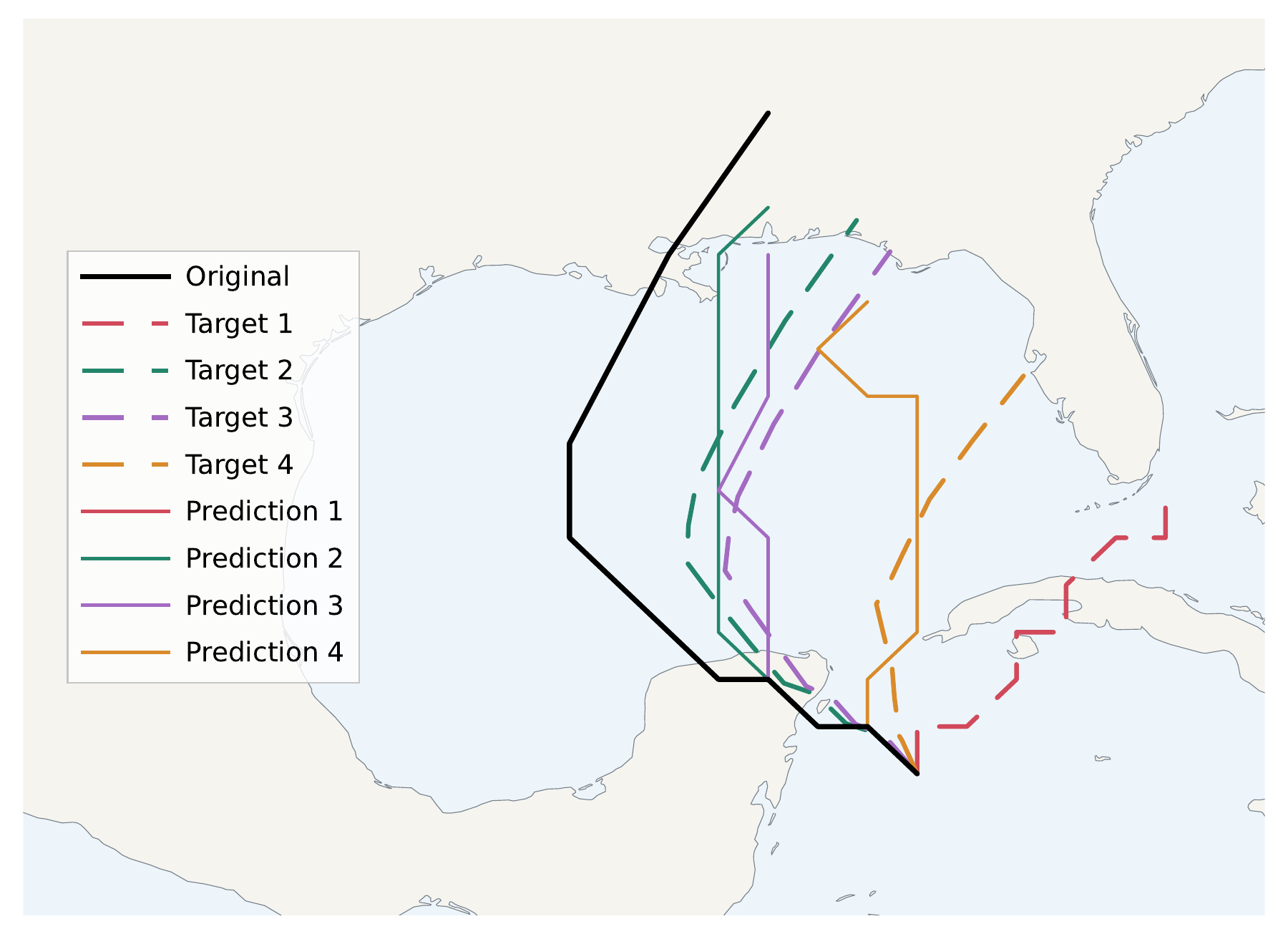}
\caption{Upstream attack visualization under varying target trajectory deviations for hurricane \textit{Zeta} (2020). Note that Target 0 corresponds to removing the original forecasted trajectory.}
\label{fig:delta_end2end}
\end{figure}

To illustrate this limitation, we use Hurricane \textit{Zeta} (2020) as a case study. Following the procedure described in Section~\ref{sec:upstream_attack}, we generate the adversarial upstream inputs, $\hat{\mathbf{X}}'$, associated with 5 different target trajectories. 
Target 0 represents the complete suppression of the original forecasted trajectory.
Targets 1 through 4 correspond to adversarial target trajectories that are progressively closer to the original forecasted track.
As shown in Figure~\ref{fig:delta_end2end}, the end-to-end \textit{Cyc-Attack} successfully suppresses the original forecasted trajectory, as evidenced by the absence of a predicted trajectory when the suppressed original track is specified as the adversarial target. Furthermore, the alignment between the end-to-end predicted trajectory and the adversarial target trajectory improves as the target moves closer to the original track.
The quantitative results for hurricane \textit{Zeta} are presented in Table~\ref{tab:delta_regional_evaluation}. For the farthest target, with a mean distance of $754.5759$ km from the original trajectory, \textit{TempestExtremes} fails to detect any TC locations in the target region from the forecasts generated by \textit{GraphCast} using the adversarial input generated by the end-to-end \textit{Cyc-Attack}, resulting in both TPR and detection rate (DR) of $0$. Although the original trajectory is completely removed, the attack fails to induce the desired target trajectory. When the target moves closer to the original trajectory, with a mean distance of $479.5773$ km, \textit{TempestExtremes} successfully detects several TC locations along the initial segment of the target, increasing the TPR to $0.4$. However, the predicted trajectory deviates substantially from the target in the later forecast steps and does not satisfy the trajectory-level detection criterion. Consequently, the DR remains $0$, despite the improved location-level accuracy.
Next, when the target moves even closer to the original trajectory, with a mean distance of only $239.7887$ km, the location-level TPR remains at $0.4$, while the trajectory DR increases to $1$, indicating that the predicted trajectory satisfies the TC detection criterion. Finally, for the closest target, with a mean distance of $179.8415$ km, the TPR further improves to $0.5$, while the DR remains at $1$. These results suggest that reducing the deviation from the original trajectory improves the attacker's ability to steer the forecast toward the desired target trajectory.

\begin{table}[tb]
\centering
\caption{Upstream attack performance under varying target trajectory deviations for hurricane \textit{Zeta} (2020). The reported results below are constrained to the region shown in Figure~\ref{fig:delta_end2end}.}
\label{tab:delta_regional_evaluation}
\small
\setlength{\tabcolsep}{5pt}
\begin{tabular}{lrrrrr}
\toprule
Target & Mean Dist. (km) & FPR$\downarrow$ & FNR$\downarrow$ & TPR$\uparrow$ & DR$\uparrow$ \\
\midrule
1 & 754.5759 & 0.0000 & 1.0000 & 0.0000 & 0 \\
2 & 479.5773 & 0.0157 & 0.6000 & 0.4000 & 0 \\
3 & 239.7887 & 0.0137 & 0.6000 & 0.4000 & 1 \\
4 & 179.8415 & 0.0118 & 0.5000 & 0.5000 & 1 \\
\bottomrule
\end{tabular}
\end{table}

\subsubsection{Complexity Analysis}

We profiled the computational cost of \textit{Cyc-Attack} on Google Colab using a single NVIDIA A100 GPU with 80 GB of memory. For the downstream attack, 1,000 optimization steps across 10 tropical cyclone cases with four adversarial target settings required 1.17 hours in total. Including \textit{TempestExtremes} post-processing, the complete downstream attack pipeline took 1.23 hours, averaging 0.031 hours per case--target pair, with peak GPU and CPU memory usage of 16.61 GiB and 6.21 GiB, respectively. The end-to-end attack incurred higher computational overhead, averaging 0.149 hours per case--target pair, with peak GPU and CPU memory usage of 38.38 GiB and 7.87 GiB, respectively. Surrogate-model pre-training, including a 30-trial Optuna hyperparameter search and final training with the selected configuration, required approximately 6 hours.

\subsubsection{Evaluation of Defense Mechanism}
\label{sec:Defending Mechanism}

{
To assess the detectability of adversarial inputs, we apply an unsupervised principal component analysis (PCA) detector on clean TC1 inputs, excluding static variables and samples within seven days of any TC2 case to prevent temporal leakage. For each input variable, the two input time steps are concatenated and standardized. We then apply PCA to the geospatio-temporal map of each variable independently in a way that retains 95\% of its respective variance. Variable-wise reconstruction errors are then computed using five-fold out-of-fold (OOF) statistics. Given a potentially manipulated input sample $\mathbf{\hat{X}}$, the anomaly score $S$ is computed from the reconstructed errors as follows,
\begin{equation}
\begin{aligned}
S_v({\mathbf{\hat{X}}})&=\frac{1}{|\mathcal{C}_v|}\sum_{k\in\mathcal{C}_v}\frac{e_k({\mathbf{\hat{X}}})}{m_k}, \
S(\mathbf{\hat{X}})=\sum_{v=1}^{11}w_vS_v({\mathbf{\hat{X}}}),
\end{aligned}
\label{eq:pca_score}
\end{equation}
where $S_v$ is the mean normalized PCA reconstruction error of variable group $v$, $\mathcal{C}_v$ denotes its channels of variable group $v$, $e_k$ is the mean squared reconstruction error of channel $k$, and $m_k$ is its mean OOF reconstruction error. The original detector assigns uniform weights $w_v=1/11$ and flags an input as anomalous when $S({\mathbf{\hat{X}}})>\theta_\alpha$, where $\theta_\alpha$ is calibrated from TC1 OOF scores at $\alpha\in\{0.95,0.98,0.99\}$. As shown in Table~\ref{tab:pca_detection}, \textit{Cyc-Attack} exhibits substantially lower detection rates than ALA.
}

{
To improve the anomaly detection, we construct a set of adversarial input samples using our end-to-end Cyc-Attack method. We then introduce an attack-aware PCA defender that retains the PCA models while learning variable-group weights from the constructed \textit{Cyc-Attack} adversarial inputs:
\begin{equation}
a_v=\frac{\max(\overline{\Delta S}_v,0)}{\sigma_v},\quad
w_v=(1-\lambda) \frac{1}{11}+ \lambda\frac{a_v}{\sum_{j=1}^{11}a_j},
\label{eq:pca_defense_weights}
\end{equation}
where $\overline{\Delta S}_v$ is the mean difference $S_v(\mathbf{X}')-S_v(\mathbf{\hat{X}})$ between paired adversarial and clean inputs, $\lambda=0.5$, and $\sigma_v$ is the standard deviation of $S_v$ on TC1 OOF samples. The weights are shrunk toward uniform weighting, with uniform weights used if all $a_v$ vanish. Detection thresholds are recalibrated using TC1 OOF scores. We employ leave-one-case-out evaluation, learning weights from nine TC2 cases and testing on Target 2 and clean inputs from the held-out case. As shown in Table~\ref{tab:pca_defense}, detection rates increase from 30\% to 80\% and from 10\% to 70\% at the 95th and 98th percentiles, respectively, with 0\% observed FPR.
}

\begin{table}[t]
\centering
\caption{{PCA-based anomaly detection results on TC2. Detection rates represent TPR for adversarial inputs.}}
\label{tab:pca_detection}
\resizebox{\columnwidth}{!}{%
\begin{tabular}{lcccc}
\toprule
\multirow{2}{*}{Input} & \multirow{2}{*}{Reconstruction Error} &
\multicolumn{3}{c}{Confidence (Anomaly Threshold)} \\
\cmidrule(lr){3-5}
& & {95\% (1.156)} & {98\% (1.424)} & {99\% (6.838)} \\
\midrule
Clean      & 0.825   & --    & --    & -- \\
ALA        & 164.884 & 100\% & 100\% & 100\% \\
Cyc-Attack & 1.120   & 30\%  & 10\%  & 0\% \\
\bottomrule
\end{tabular}%
}
\end{table}

\begin{table}[tb]
\centering
\small
\caption{{Comparison of original PCA detection and attack-aware PCA defense against \textit{Cyc-Attack} Target 2.}}
\label{tab:pca_defense}
\begin{tabular}{lccc}
\toprule
{Detector} & {95\%} & {98\%} & {99\%} \\
\midrule
Original PCA & 30\% & 10\% & 0\% \\
Attack-Aware PCA & 80\% & 70\% & 0\% \\
\bottomrule
\end{tabular}%
\end{table}

\section{Limitations and Ethics Statement}
\label{sec:Limitations and Ethics}

{
Our experiments reveal a limitation of upstream attacks by \textit{Cyc-Attack} when the target trajectory deviates substantially from the original TC track. Although downstream attacks by \textit{Cyc-Attack} can successfully achieve large trajectory deviations (see Table~\ref{tab:tc1_tc2_results} and Figure~\ref{fig:comparison_of_different_methods}), upstream attacks become more challenging as the target distance increases (see Table~\ref{tab:cyc_attack_ala_comparison} and Figure~\ref{fig:delta_end2end}). One possible explanation is that upstream attacks modify only two input time steps and rely on the weather forecasting model to propagate these perturbations through dozens of forecast steps. Steering a TC toward a distant target within a longer forecast horizon may therefore require more extensive and complex modifications to the evolving atmospheric fields, which are difficult to achieve under bounded perturbations and may compromise physical plausibility.
}

\textbf{Our work does not target a specific operational system.} The goal is to highlight potential vulnerabilities and motivate the development of more robust DLWF models as well as security-aware practices across the forecasting pipeline. {
Our experiments in Section~\ref{sec:Defending Mechanism} further demonstrate that training an attack-aware detector using adversarial input samples generated by \textit{Cyc-Attack} can substantially improve its ability to detect previously unseen attacks of the same type, highlighting the potential of adversarial examples for strengthening defensive mechanisms.
}

\section{Conclusions}
\label{sec:conclusions}

This paper presents {\textit{Cyc-Attack}}, a novel method for manipulating weather forecasts generated by DLWF models to alter the projected trajectories of TC detection systems.  
{\textit{Cyc-Attack}} trains a differentiable surrogate model to enable gradient-based attacks, incorporates a skewness-aware loss with kernel dilation to address class imbalance, and applies distance-based weighting with regularization to ensure perturbations remain realistic and imperceptible. These design choices overcome key limitations of standard attack methods, 
enabling {\textit{Cyc-Attack}} to outperform various baselines in terms of producing the desired adversarial trajectory while producing stealthier perturbations. 

We further extend \textit{Cyc-Attack} to an end-to-end attack setting by leveraging its adversarial upstream forecasts as guidance for training an end-to-end model. Specifically, these forecasts provide a reference for the desired changes in the predicted atmospheric states, enabling the model to learn adversarial perturbations to the original meteorological inputs that propagate through the weather forecasting and TC tracking pipeline, ultimately steering the predicted TC trajectories toward attacker-specified targets.

The results of this work demonstrate the feasibility of adversarial attacks on DLWF systems that steer downstream TC trajectory predictions toward an attacker’s chosen target. 

\section*{Declaration of Generative AI Usage}
The authors used ChatGPT (OpenAI) solely as an auxiliary tool to support the research process. Specifically, it was used to: (1) improve grammatical correctness and writing clarity; (2) assist in identifying relevant literature and reviewing related work descriptions for accuracy and fairness; (3) examine code for potential errors and consistency with the proposed methods; and (4) identify potential issues in mathematical derivations and experimental designs. The research contributions, methodological decisions, and conclusions were developed by the authors. All AI-assisted suggestions were critically reviewed, and any adopted changes were independently verified by the authors, who take full responsibility for the content and findings of this work.

\section*{Acknowledgments}
{
This research is supported by the U.S. National Science Foundation under grant IIS-2453100. Any use of trade, firm, or product names is for descriptive purposes only and does not imply endorsement by the U.S. 
}

\bibliography{yue}
\bibliographystyle{IEEEtran}

\clearpage
\section*{Supplementary Materials}

\subsection*{A.TempestExtremes Software for TC trajectory prediction}

\textit{TempestExtremes}~\cite{ullrich2021tempestextremes}, a black-box TC trajectory detection software, operates in two stages: (1) the first stage is to identify candidate TC locations as local minima in \textit{mean sea level pressure} (MSLP) that are not within $6^{\circ}$ of a deeper MSLP minimum, enclosed by a 200 hPa closed contour within a $5.5^{\circ}$ radius; and co-located with maxima in the \textit{geopotential thickness} field computed as the difference in geopotential height between 300 and 500 hPa (Z300–Z500), where the maxima are enclosed by a $58.8$ m\textsuperscript{2}s\textsuperscript{–2} closed contour within a $6.5^{\circ}$ radius, allowing stronger peaks within $1.0^{\circ}$. The variables associated with each candidate location, including \textit{mean sea level pressure}, \textit{elevation}, and the maximum regional \textit{10-meter wind speed}, are also the output of the first stage.
(2) The second stage is to stitch these candidate locations into trajectories constrained by a maximum $8^{\circ}$ step distance, a maximum $24$-hour gap, a minimum $54$-hour lifetime, and at least 10 time steps with \textit{10-meter wind speed} no less than 10 m/s, \textit{elevation} not exceeding 150 m, and \textit{latitude} between $-50^{\circ}$ and $50^{\circ}$. 

In short, the input variables to \textit{TempestExtremes} includes \textit{mean sea level pressure}, \textit{10-meter wind speed}, \textit{elevation}, and \textit{geopotential thickness}, and the output consists of TC trajectories obtained by merging the candidate locations within a time period. 

\subsection*{B.Datasets and Preprocessing}

Three pre-trained checkpoints of the \textit{GraphCast} model~\cite{lam2023learning} have been released by Google DeepMind. 
They differ in data sources (\textit{ERA5} or \textit{ERA5+HRES}), spatial resolution ($0.25^{\circ}$ or $1.0^{\circ}$), the number of pressure levels ($37$ or $13$), mesh configurations (\textit{2to6} or \textit{2to5}), and whether \textit{precipitation} is used only as an output or as both input and output. Given the model-agnostic nature of {\textit{Cyc-Attack}}, it is not constrained by the specific architecture of the DLWF model. 
In this study, we used the checkpoint trained on \textit{ERA5}, with $1.0^{\circ}$ resolution, $13$ pressure levels, mesh \textit{2to5}, and precipitation included as both input and output\footnote{The selected checkpoint is \textit{"GraphCast\_small - ERA5 1979-2015 - resolution 1.0 - pressure levels 13 - mesh 2to5 - precipitation input and output.npz"}.}.


The output data of the \textit{GraphCast} model includes the five single-level variables and six multi-level variables described above. All data are provided at a $6$-hour temporal resolution, corresponding to 00Z, 06Z, 12Z, and 18Z each day, with $2$ time steps used as input to the \textit{GraphCast} model, and $1$, $4$, $12$, $20$, or $50$ time steps used as forecast outputs. 
The input and output data are not standardized, as there is a built-in standardization process within the \textit{GraphCast} model. Additionally, the hourly precipitation values in ERA5 data are aggregated over six-hour windows.


As previously noted, the input to \textit{TempestExtremes}~\cite{ullrich2021tempestextremes} are \textit{mean sea level pressure}, \textit{10-meter wind speed}, \textit{elevation}, and \textit{geopotential thickness}. Among these variables, \textit{wind speed} and \textit{geopotential thickness} are not directly available in the \textit{GraphCast} forecast output, but can be derived. \textit{Wind speed} as the magnitude of the u- and v-component of the wind vector, while \textit{geopotential thickness} is obtained by taking the difference between the geopotential heights at 300 hPa and 500 hPa (i.e., Z300-Z500). Additionally, \textit{elevation} can be approximated by dividing the surface geopotential by 9.80665, where the surface geopotential is obtained from external datasets.

\subsection*{C.Adversarial Target Trajectory Construction}
We propose a strategy for constructing adversarial downstream targets. Given the original downstream prediction $\hat{\mathbf{Z}} \in \{0,1\}^{\beta \times r \times c}$, we represent each trajectory as a sequential path $\hat{\mathbf{p}} = \{\hat{\mathbf{p}}_\tau \in \mathbb{R}^2 \mid \tau = 1, \dots, \beta \}$, where $\hat{\mathbf{p}}_\tau = (\lambda_\tau, \phi_\tau)$ denotes the longitude and latitude (in degrees) at time step $\tau$. Our goal is to construct an adversarial trajectory $\hat{\mathbf{p}}' = \{ \hat{\mathbf{p}}'_\tau \in \mathbb{R}^2 \mid \tau = 1, \dots, \beta \}$ that preserves the temporal step length of the corresponding original one while progressively deviating in direction. The constructed adversarial trajectory becomes the adversarial downstream targets $\mathbf{\hat{Z}}^* \in \{0,1\}^{\beta \times r \times c}$.

In detail, the adversarial trajectory $\hat{\mathbf{p}}'$ begins at the same origin as the original trajectory $\hat{\mathbf{p}}$, i.e., $\hat{\mathbf{p}}'_1 = \hat{\mathbf{p}}_1$. For each subsequent step $\tau > 1$, we compute the great-circle distance $\hat{d}_\tau$ between $\hat{\mathbf{p}}_{\tau-1}$ and $\hat{\mathbf{p}}_{\tau}$ using the spherical law of cosines:
\begin{align}
\small
&\cos \Delta \sigma_\tau = \sin \Phi_{\tau-1} \sin \Phi_\tau + \cos \Phi_{\tau-1} \cos \Phi_\tau \cos(\Lambda_\tau - \Lambda_{\tau-1}), \nonumber\\
&\hat{d}_\tau = R \cdot \arccos\big(\operatorname{min}(\max(\cos \Delta \sigma_\tau, -1), 1)\big),\nonumber
\end{align}
where $(\Lambda_\tau, \Phi_\tau)$ and $(\Lambda_{\tau-1}, \Phi_{\tau-1})$ are the radian representations of the coordinates, and $R$ is the Earth's radius in kilometers.

Let $\mathcal{M} = \{ \boldsymbol{\delta}_j \in \mathbb{R}^2 \mid j = 1, \dots, 8 \}$ denote the set of 8-connected compass direction unit vectors on the 2D plane. Each direction $\boldsymbol{\delta}_j$ corresponds to a compass bearing $\theta_j \in [0^\circ, 360^\circ)$. At each time step, we sample a direction index $j^*$ according to a probability distribution derived from angular deviation from the original path.

To quantify directional deviation, we first define the cosine similarity between the original displacement vector and each candidate direction. Let $\hat{\mathbf{v}}_\tau = \hat{\mathbf{p}}_\tau - \hat{\mathbf{p}}_{\tau-1}$ be the original vector, and let $\hat{\mathbf{v}}'_{\tau - 1}=\hat{\mathbf{p}}'_{\tau-1} - \hat{\mathbf{p}}'_{\tau-2}$ be the previous adversarial displacement when $\tau > 2$. We compute
\[
\cos \theta_j^{\text{orig}} = \frac{\hat{\mathbf{v}}_\tau^\top \boldsymbol{\delta}_j}{\|\hat{\mathbf{v}}_\tau\|}, \quad
\cos \theta_j^{\text{adv}} = \frac{{\hat{\mathbf{v}}'^{\top}_{\tau - 1}} \boldsymbol{\delta}_j}{\|{\hat{\mathbf{v}}'_{\tau - 1}}\|},
\]
which measures the deviation magnitude from the current original step and the coherence with the prior adversarial step, respectively. Then, we assign a score $s_j$ to each direction $\delta_j$ by
\[
s_j = \gamma_1 \cdot \exp(- \cos \theta_j^{\text{orig}}) + \gamma_2 \cdot \exp(\cos \theta_j^{\text{adv}}).
\]
These scores are normalized into probabilities $P_j = s_j / \sum_{\ell=1}^8 s_\ell$, and the direction $j^*$ with the highest probability is selected.

To ensure smooth and plausible deviation, we compute a blended unit direction vector by summing the original direction $\hat{\mathbf{v}}_\tau$ and the selected direction $\boldsymbol{\delta}_{j^*}$ by
\[
\mathbf{u}_\tau = \frac{\hat{\mathbf{v}}_\tau + \boldsymbol{\delta}_{j^*}}{\|\hat{\mathbf{v}}_\tau + \boldsymbol{\delta}_{j^*}\|},
\]
and then convert it to a compass bearing $\theta_\tau$ by
\[
\theta_\tau = \operatorname{atan2}(u_{\tau, 0}, u_{\tau, 1}),
\]
converted to degrees in $[0^\circ, 360^\circ)$.

The next adversarial trajectory location is obtained by projecting from the previous position $\hat{\mathbf{p}}'_{\tau-1} = (\lambda_{\tau-1}', \phi_{\tau-1}’)$ a geodesic distance $\hat{d}_\tau$ along bearing $\theta_\tau$, using the following great-circle formula:
\begin{align}
\phi_{\tau}' &= \arcsin\left(\sin\phi_{\tau-1}'\cos\frac{\hat{d}_\tau}{R} + \cos\phi_{\tau-1}'\sin\frac{\hat{d}_\tau}{R}\cos\theta_\tau\right), \nonumber\\
\lambda_{\tau}' &= \lambda_{\tau-1}' + \operatorname{atan2}\Bigl(\sin\theta_\tau\sin\frac{\hat{d}_\tau}{R}\cos\phi_{\tau-1}', \nonumber\\
&\qquad \cos\frac{\hat{d}_\tau}{R} - \sin\phi_{\tau-1}'\sin\phi_{\tau}'\Bigr).\nonumber
\end{align}
where all angles are in radians and $R$ is the Earth's radius in kilometers. By iterating this process from $\tau = 2$ to $\beta$, we obtain an adversarial trajectory $\hat{\mathbf{p}}' = \{ \hat{\mathbf{p}}'_\tau \in \mathbb{R}^2 \mid \tau = 1, \dots, \beta \}$ that preserves geodesic step lengths while progressively deviating in direction from the original path $\hat{\mathbf{p}} = \{ \hat{\mathbf{p}}_\tau \in \mathbb{R}^2 \mid \tau = 1, \dots, \beta \}$.

For each original downstream output of \emph{TempestExtremes} $\mathbf{\hat{Z}}$, if it contains at least one trajectory, it will be included in our pool of valid samples for constructing adversarial targets. Otherwise, it will be discarded.  
As a result, among all $285$ test samples, we retain only $149$ samples containing detected TC trajectories. If a sample contains $N$ detected TC trajectories, we construct $N$ adversarial samples, targeting one TC trajectory at a time, while keeping the other $N{-}1$ trajectories unchanged.  
This yields $N$ adversarial versions per valid sample, with one adversarial trajectory constructed in each.  

Finally, we construct each adversarial downstream $\mathbf{\hat{Z}}^* \in \mathbb{R}^{\beta \times r \times c}$ from the original one $\mathbf{\hat{Z}} \in \mathbb{R}^{\beta \times r \times c}$ by replacing the trajectory locations along the original predicted trajectory with those along the constructed adversarial trajectory, while leaving all others unchanged.

\subsection*{D.Pre-training the Surrogate Model}

Serving as a proxy for the black-box \textit{TempestExtremes} software, the surrogate model's input data are the same as that of \textit{TempestExtremes}, i.e., \textit{mean sea level pressure}, \textit{geopotential thickness}, \textit{10-meter wind speed}, and \textit{elevation}, extracted from the constructed ground truth ERA5 dataset across $12$ time steps within the forecast window of the \textit{GraphCast} model. All training inputs are standardized using the variable-wise mean and standard deviation computed from our collected dataset. The output corresponds to a dilated region, with radius $R_S$, centered at the trajectory locations detected by the black-box \textit{TempestExtremes} software at each of the $12$ time steps.

We use the \textit{Adam} optimizer~\cite{kingma2014adam}, and perform hyperparameter tuning using \textit{Optuna}~\cite{akiba2019optuna}, 
an automatic optimization framework. A total of $30$ trials are conducted to search for optimal values of the following hyperparameters: \textit{learning rate} in the range $[1\text{e}{-5}, 1\text{e}{-2}]$, $\beta_1$ in $[0.5, 0.99]$, $\beta_2$ in $[0.9, 0.9999]$, \textit{weight decay} in $[1\text{e}^{-6}, 1\text{e}^{-3}]$, \textit{convolution stride size} from the set $\{1, 2, 3, 4, 6, 8\}$, and \textit{number of epochs} in the range $[10, 100]$ for training the \textit{DeepLabV3+} model. For each trial, the model achieving the best validation performance is retained. The optimal hyperparameters—\textit{learning rate} = $5.92 \times 10^{-4}$, $\beta_1 = 0.9101$, $\beta_2 = 0.9119$, weight decay = $6.48 \times 10^{-6}$, \textit{stride size} = $1$, and \textit{number of epochs} = $77$—were identified in trial 1, yielding a best validation loss of $16.72$.

Unlike \textit{TempestExtremes}, the surrogate model does not output additional variables associated with each detected TC location, such as \textit{mean sea level pressure}, \textit{elevation}, and the regional maximum \textit{10-meter wind speed}.
Instead, for each location detected by the surrogate model, the variables \textit{mean sea level pressure} and \textit{elevation} are taken directly from the corresponding weather data at that location. To compute the maximum regional \textit{10-meter wind speed}, let $\lambda_{i_0j_0}$ and $\phi_{i_0j_0}$ denote the \textit{longitude} and \textit{latitude}, respectively, of a candidate location $(i_0,j_0)$ (in degrees) while $w_{i_0j_0}$ denote the \textit{10-meter wind speed} at the location. The value at a candidate location is considered missing if $w_{i_0j_0} = \operatorname{NaN}$ or if $|w_{i_0j_0} - 10^{20}| < 10^{-6}$. The maximum regional \textit{10-meter wind speed} is constructed as follows. (i) For each candidate location $(i_0,j_0)$, map its coordinates $(\lambda_{i_0j_0}, \phi_{i_0j_0})$ to $(\Lambda_{i_0j_0}, \Phi_{i_0j_0})$ in radians as 
$
\Lambda_{i_0j_0} = \left( \frac{\lambda_{i_0j_0}}{180} \pi \right) \bmod (2\pi), \quad \Phi_{i_0j_0} = \frac{\phi_{i_0j_0}}{180} \pi,
$
where $\Lambda_{i_0j_0} \in [0, 2\pi)$.
(ii) For each candidate location $(i_0,j_0)$, compute the central angle $\Delta \sigma_{ij}$ in radians between $(i_0,j_0)$ and any point $(i,j)$ as
$
\Delta \sigma_{ij} = \arccos (c_{ij}),
$
where
$
c_{ij} = \sin \Phi_{i_0j_0} \sin \Phi_{ij} + \cos \Phi_{i_0j_0} \cos \Phi_{ij} \cos (\Lambda_{ij} - \Lambda_{i_0j_0}),
$
with $c_{ij}$ clipped to the interval $[-1,1]$ such that $\Delta \sigma_{ij}$ is clipped to the interval $[0,\pi]$. The corresponding great-circle distance between $(i,j)$ and $(i_0,j_0)$ is then computed as
$
D_{ij} = \frac{180}{\pi} \Delta \sigma_{ij}.
$
(iii) For the candidate location \((i_0,j_0)\), define the set of neighboring nodes located along eight directions (N, NE, E, SE, S, SW, W, NW) and within a great-circle distance of \(2\) as
$
\mathcal{S}_{(i_0,j_0)} = \left\{ (i,j) \ \middle| \ D_{ij} \leq 2;\ w_{ij} \text{ is not missing} \right\}.
$
(iv) If $\mathcal{S}_{(i_0,j_0)}$ is nonempty, define the maximum regional \textit{10-meter wind speed} at candidate location $(i_0,j_0)$ as $\max_{(i,j) \in \mathcal{S}_{(i_0,j_0)}} w_{ij}$; otherwise, it is defined as $w_{i_0j_0}$. 

Note that, due to subtle differences in floating-point precision between Python (used in the reconstruction process above) and C++ (used in \textit{TempestExtremes}), some discrepancies may arise near the threshold distance $D_{ij} = 2.0$. Specifically, some neighbors identified by \textit{TempestExtremes} as within $D_{ij}= 2.0 - \epsilon$ may be computed as $D_{ij} = 2.0 + \epsilon$ in Python, and thus excluded, and vice versa, where $\epsilon$ is a very small error. To resolve this, we uniformly set the threshold to $D_{ij} = 2.0 + \epsilon$ with $\epsilon = 10^{-8}$ in both implementations. This ensures that the computed maximum regional \textit{10-meter wind speed} exactly matches each from \textit{TempestExtremes}.

\subsection*{E.Comparison of Adversarial Attack Methods}

\begin{figure*}[tb]
    \centering
    \includegraphics[width=1.0\textwidth]{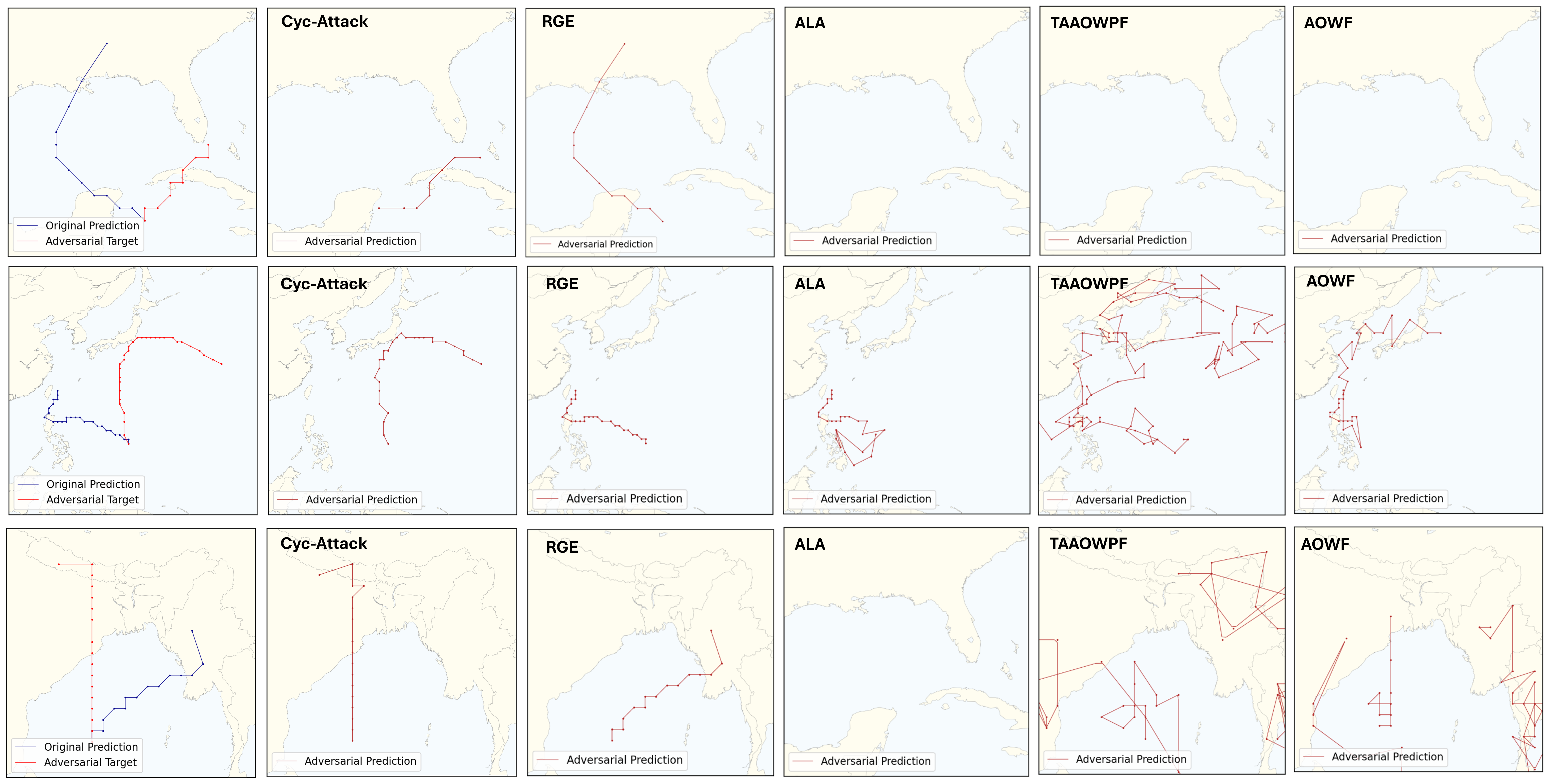}  
    \caption{Downstream attack performance of \textit{Cyc-Attack} and baseline methods on Hurricane \textit{Zeta} (2020), typhoon \textit{Megi} (2010), and typhoon \textit{Nargis} (2008), arranged from top to bottom. The visualization follows the conventions in Figure~\ref{fig:comparison_of_different_methods}.}
\label{fig:comparison_of_different_methods_more}
\end{figure*}

Figure~\ref{fig:comparison_of_different_methods_more} provides supplementary visualizations for comparing the performance of different adversarial attack methods, which aligns with the main text: {\textit{Cyc-Attack}} attains a higher detection rate (TPR), more accurately shifts predictions to the adversarial target while suppressing the original trajectory, and yields fewer false alarms (lower FPR).

\subsection*{F.Sensitivity Analysis}

\begin{figure}[tb]
    \centering
    \includegraphics[width=1.0\linewidth]{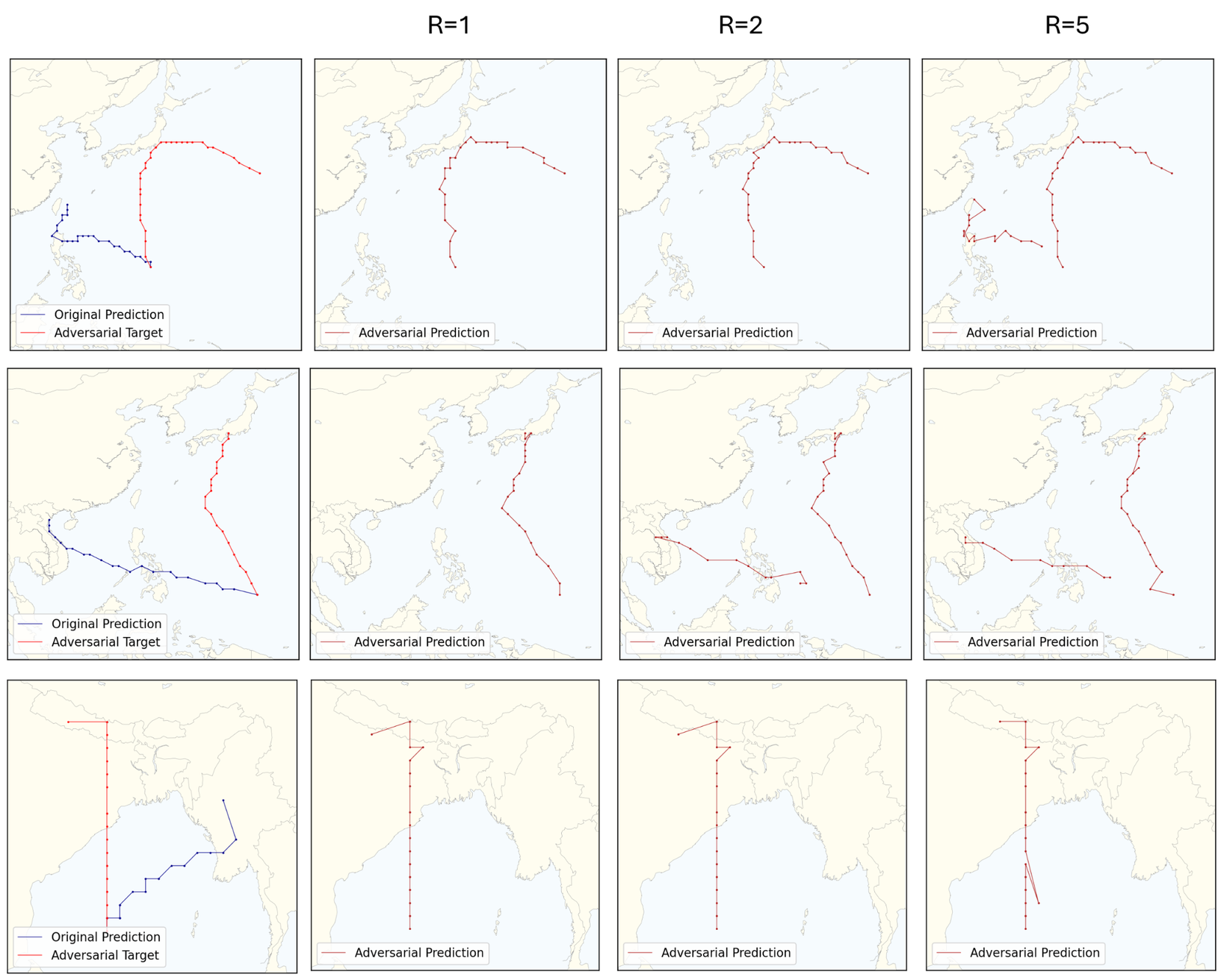}  
    \caption{
    Downstream attack performance of \textit{Cyc-Attack} under different dilation radii for typhoon \textit{Megi} (2010), typhoon \textit{Haiyan} (2013), and typhoon \textit{Nargis} (2008), arranged from top to bottom. The visualization follows the conventions in Figure~\ref{fig:adv_attack_different_radius}.}
\label{fig:supplemenraty_adv_attack_different_radius}
\end{figure}

Figure~\ref{fig:supplemenraty_adv_attack_different_radius} provides supplementary results illustrating the sensitivity of {\textit{Cyc-Attack}} to the choice of dilation radius. These results are consistent with the conclusions presented in the main text: as the dilation radius increases, the generated adversarial predicted trajectories exhibit more pronounced zigzagging, and, to a certain extent, induce an increased number of false-positive trajectories. 

\end{document}